%% file: main.tex
\documentclass[runningheads]{llncs}

\usepackage[final,year=2026,ID=2042]{eccv}
\usepackage{eccv}

\usepackage{eccvabbrv}

\usepackage{graphicx}
\usepackage{multirow}
\usepackage[utf8]{inputenc} 
\usepackage[T1]{fontenc}    
\usepackage{url}            
\usepackage{booktabs}       
\usepackage{amsfonts}       
\usepackage{nicefrac}       
\usepackage{microtype}      
\usepackage{adjustbox}
\usepackage{pifont}
\usepackage{bbding}
\usepackage{amsmath}
\usepackage{makecell}
\usepackage{tabularx}
\usepackage{wrapfig}
\usepackage{caption}

\usepackage{listings} 
\usepackage{tikz}
\usepackage{enumitem}
\usepackage{colortbl}
\definecolor{darkblue}{RGB}{94,110,186}
\definecolor{Gray}{gray}{0.5}

\definecolor{darkGreen}{RGB}{92, 148, 110}
\definecolor{asparagus}{RGB}{0.53, 0.66, 0.42}
\newcommand{\darkGreen}[1]{\textcolor{darkGreen}{#1}}

\newcommand{\model}{\textbf{Self-Improving Demonstrations}}
\usepackage[accsupp]{axessibility}  

\usepackage[pagebackref,breaklinks,colorlinks,citecolor=eccvblue]{hyperref}
\usepackage{orcidlink}

\begin{document}

\title{Learning Goal-Oriented Vision-and-Language\\ Navigation with Self-Improving\\ Demonstrations at Scale} 

\titlerunning{ }

\authorrunning{ }
\author{Songze Li$^{*1,3}$  Zun Wang$^{*\spadesuit2}$  Gengze Zhou$^{4}$  Jialu Li$^{2}$  Xiangyu Zeng$^{1,5}$\\ 
Ziyang Gong$^{6}$  Limin Wang$^{1,5}$  Yu Qiao$^{1}$  Qi Wu$^{4}$  Mohit Bansal$^{2}$  Yi Wang$^{
1}$\vspace{0.2em}\\ 
$^1$Shanghai AI Laboratory  $^2$UNC Chapel Hill  $^3$Fudan University \\
$^4$The University of Adelaide  $^5$Nanjing University  $^6$Shanghai Jiaotong University
    \vspace{0.2em}\\
{\tt \href{https://github.com/OpenGVLab/SID-VLN}{https://github.com/OpenGVLab/SID-VLN}}
}


\institute{}

\maketitle

\begingroup
\renewcommand\thefootnote{}\footnotetext{*Equal contribution. $^{\spadesuit}$Project lead.}%
\addtocounter{footnote}{-1}%
\endgroup

\begin{figure}[ht]
  \vspace{-20.0pt}
  \centering
    \includegraphics[width=0.85\textwidth]{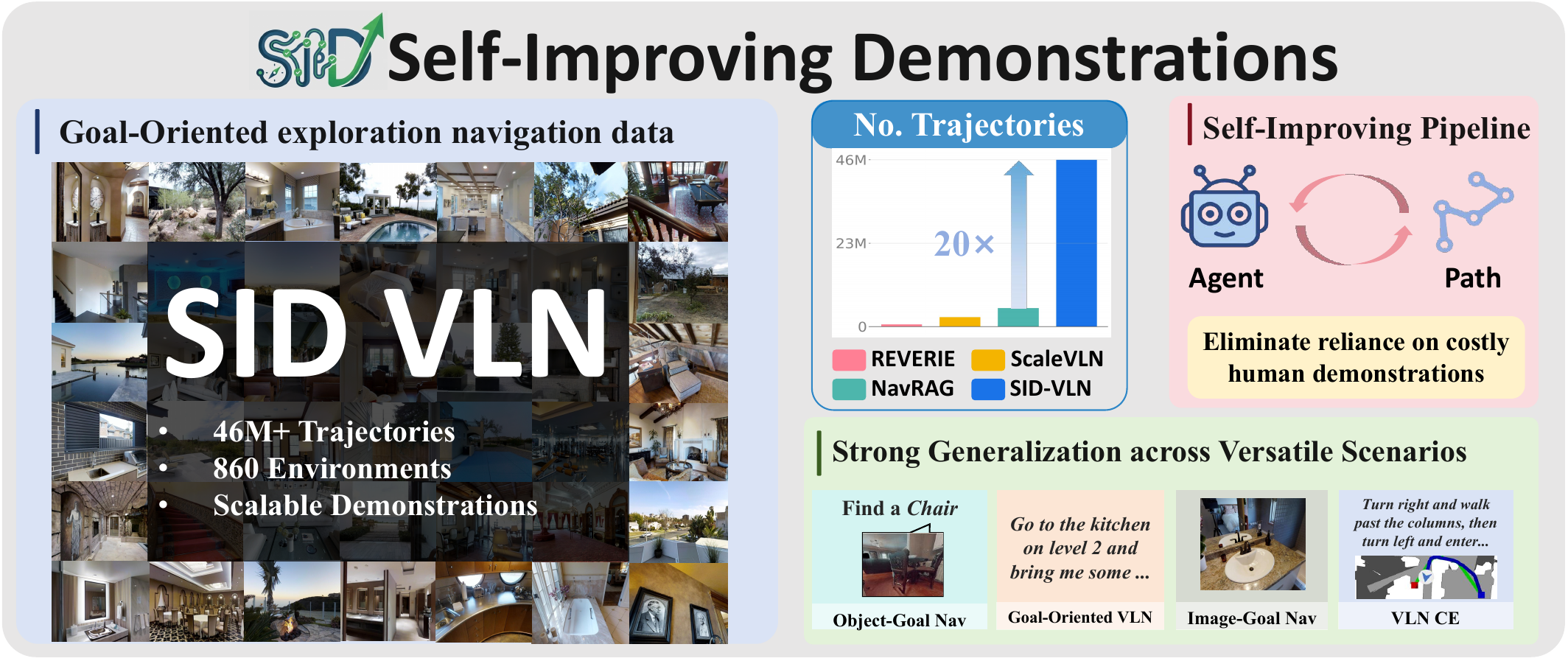}
  \caption{\textbf{SID-VLN} generates over 46M self-improving exploration trajectories across 860 environments, exhibiting strong generalization across diverse goal modalities.}
  \label{fig:teaser}
  \vspace{-30.0pt}
\end{figure}

\begin{abstract}
    Goal-oriented vision-language navigation requires robust exploration capabilities for agents to navigate to specified goals in unknown environments without step-by-step instructions. Existing methods tend to exclusively utilize shortest-path trajectories, lacking effective exploration priors for training navigation agents. To address the above challenges, we present SID, a goal-oriented vision-and-language navigation learning approach with Self-Improving Demonstrations. Specifically, SID learns an initial agent on the shortest-path data sampled from environments and then leverages this agent to generate novel exploration trajectories. The novel rollouts provide demonstrations with stronger exploration strategies to train a better agent, which in turn produces higher-quality agent demonstrations for the next round of training. We show that this iterative self-improving pipeline readily scales to new environments, and the resulting demonstrations are highly transferable, elevating the performance ceiling across a variety of vision-and-language navigation tasks. Extensive experiments demonstrate that SID significantly boosts the exploration capabilities and generalization of navigation agents. The resulting agent achieves new state-of-the-art performance on goal-oriented vision-and-language navigation benchmarks, including REVERIE, SOON as well as strong transferability to object-goal navigation and VLN-CE. It notably achieves a 50.9\% success rate on the unseen validation splits of SOON, surpassing prior leading approaches by a margin of 13.9\%.
\end{abstract}

\section{Introduction}
\label{sec:intro}
Developing embodied agents capable of following human instructions to navigate the real world has been a longstanding goal in artificial intelligence. Research in this area encompasses various tasks, including Vision-and-Language Navigation (VLN) with step-by-step instructions~\cite{zhang2024narrowing} (\textit{e.g.}, R2R~\cite{anderson2018r2r}, RxR~\cite{anderson2020rxr}) and goal-oriented VLN tasks~\cite{chaplot2020object} (\textit{e.g.}, REVERIE~\cite{qi2020reverie} and SOON~\cite{zhu2021soon}). While step-by-step instructions like ``\textit{Go down the hallway, turn left at the kitchen, and stop in front of the fridge}'' give detailed path descriptions, they are impractical for real-world applications where humans prefer concise commands, using Goal-oriented instructions like ``\textit{bring me a spoon from the kitchen}''. 

Given that goal-oriented instructions primarily depict the target and its surroundings, agents need to explore the environment to find it. Early efforts either utilize Reinforcement Learning (RL) for exploration~\cite{mirowski2016learning,zhu2017target,mousavian2019visual,gupta2017cognitive}, or learn a navigation policy from human demonstrations~\cite{ramrakhya2022habweb}.
The former requires sophisticated reward engineering, suffering from unstable training and massive compute demands~\cite{wijmans2020ddppo,krantz2021waypoint}, while the latter leans heavily on costly human annotations that are hard to collect and transfer to vision-language navigation tasks. Recently, with improved models ~\cite{vaswani2017attention} and large-scale synthetic instructions~\cite{chen2022hm3dlearning, wang2023scalevln,li2024panogen, wang2024bootstrapping, wang2025navrag, wang2025panogenplus}, a number of methods~\cite{chen2022duet,wang2023gridmm,Wang2024GOAT} leverage pure Imitation Learning (IL) to develop exploration policies. Despite these progresses, these models are typically exclusively pretrained by greedily imitating the shortest-path samples, leading to agents devoid of effective exploration priors. Although online finetuning aids in error correction during decision making, this exploration-agnostic pretraining inherently limits generalization in unseen environments.

\begin{figure*}[t]
    \vspace{-2.0pt}
    \centering
    \begin{subfigure}[b]{0.325\textwidth}
        \centering
        \includegraphics[width=0.7\textwidth]{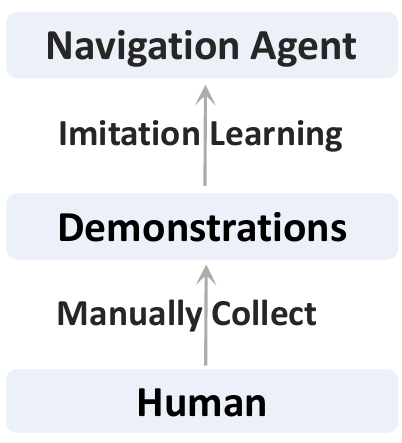}
        \caption{\resizebox{0.89\linewidth}{!}{IL on Human Demonstrations.}}
        \label{fig:IL_on_human}
    \end{subfigure}
    \hfill
    \begin{subfigure}[b]{0.325\textwidth}
        \centering
        \includegraphics[width=0.7\textwidth]{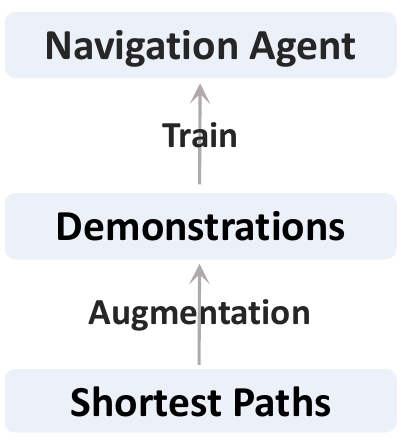}
        \caption{\resizebox{0.88\linewidth}{!}{Shortest-Path Augmentation.}}
        \label{fig:Learn_from_flywheel}
    \end{subfigure}
    \hfill
    \begin{subfigure}[b]{0.325\textwidth}
        \centering
        \includegraphics[width=0.7\textwidth]{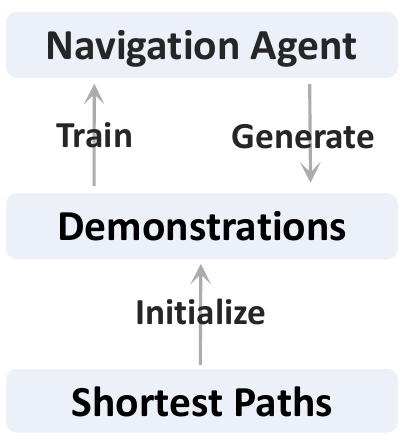}
        \caption{\resizebox{0.89\linewidth}{!}{Self-Improving Demonstrations.}}
        \label{fig:SID}
    \end{subfigure}
    \caption{\textbf{Comparison of three goal-oriented VLN training paradigms.} Learning exploration from imitating human demonstrations is costly and difficult to scale up, while learning general navigation from large-scale instruction augmentation on shortest-paths lacks exploration demonstrations. In contrast, \textbf{SID-VLN provides large-scale demonstrations on exploration strategies in an iterative self-improving approach, eliminating the reliance on costly human demonstrations.}}
    \label{fig:comparison}
    \vspace{-10.0pt}
\end{figure*}

As depicted in Figure~\ref{fig:teaser}, we propose learning from \textbf{S}elf-\textbf{I}mproving \textbf{D}emonstrations (\textbf{SID}), a goal-oriented navigation training framework that leverages the agent’s own successful exploration data as IL demonstrations, as shown in Figure~\ref{fig:comparison}. We start by applying SID to a classical goal-oriented navigation task, image-goal navigation, where data is easily acquired as each target image and starting viewpoint pair naturally forms a training sample, in base environments such as Matterport3D. Specifically, SID starts with training a base agent with shortest paths demonstrations in MP3D, then this agent infers new exploration trajectories (\textit{e.g.}, explores different rooms and distinguishes fine-grained visual elements), and only successful ones are collected as self-demonstrations to train a more capable agent. 
The agent in the new iteration is pretrained on these self-demonstrations, followed by finetuning with both self-demonstrations and shortest-path data to effectively balance exploration and exploitation. The finetuned agent generates new self-demonstrations for training the next-iteration agent.
This iterative cycle brings consistent performance gains, where improved agents generate superior exploration trajectories and, in turn enhance subsequent agent training.

When the agent's performance saturates within the initial environments after several iterations of self-demonstration learning, we extend SID by incorporating additional environments (\textit{e.g.} HM3D~\cite{ramakrishnan2021hm3d}) to expand the training corpus and continue the iterative SID training, resulting in a proficient navigation agent along with large-scale image-goal demonstration trajectories containing rich exploration priors. We then explore transferring SID trajectories to goal-oriented VLN tasks such as REVERIE and SOON. To address the scarcity of paired instructions for each target image, we propose a simple but effective augmentation method that leverages Vision-Language Models (VLMs) to generate detailed captions as instructions, producing over 46M language-goal trajectories(20$\times$ scale-up compared to prior datasets~\cite{wang2025navrag,wang2023scalevln,chen2022hm3dlearning}). The agent is first pretrained on the resulting training corpus and then finetuned on downstream tasks.

Extensive experiments show that the agent's self-demonstrations not only consistently outperform shortest-path demonstrations on image-goal navigation and VLN tasks but also benefit both pretraining and finetuning. The agent improves steadily with higher-quality self-demonstrations generated by the stronger agent at each iteration, which can be continued by introducing more environments. With our final SID-VLN data, a simple navigation agent such as DUET~\cite{chen2022duet} achieves state-of-the-art performance on goal-oriented VLN tasks, including REVERIE and SOON. Notably, it reaches a 50.9\% success rate on SOON, exceeding the previous approaches by a significant margin of 13.9\%. We also empirically verify the strong transferability of SID-VLN trajectories on object-goal navigation and VLN-CE~\cite{krantz2020beyond}. Our contributions can be listed as follows:
\vspace{-1.0pt}
\begin{itemize}[leftmargin=*]
\item We present \textbf{Self-Improving Demonstrations (SID)}, a novel iterative approach that enables a navigation agent to learn robust exploration strategies with demonstrations from its own successful trajectories, averting the dependence on costly human annotations for exploration data. 

\item We provide large-scale exploratory language-goal trajectories along with a capable navigation agent that can be used to generate additional exploration paths, which is the first to offer transferable large-scale demonstrations on exploration strategies, filling a critical gap in goal-oriented navigation.

\item We demonstrate SID's scalability across diverse environments and transferability to goal-oriented vision-and-language navigation tasks through vision-language-aligned pretraining, achieving state-of-the-art results on challenging goal-oriented VLN tasks including SOON, REVERIE. SID-VLN also showcases strong transferability to object-goal navigation and VLN-CE.
\end{itemize}

\section{Related Work}
\label{sec:formatting}
\noindent\textbf{Vision-and-Language Navigation} requires the agents to autonomously navigate in new environments by comprehending and executing natural language instructions. Currently, numerous scenarios have been proposed for VLN tasks, including navigation via detailed instructions of sequences of actions~\cite{anderson2018r2r, anderson2020rxr, thomason2020vision}, remote object navigation with coarse-grained high-level instructions~\cite{zhu2021soon, qi2020reverie}, and object-goal navigation in continuous environment ~\cite{kolve2017ai2, krantz2020beyond, xia2018gibson, hong2022bridging}. To solve this problem, previous methods concentrate primarily on LLM planning for VLN~\cite{zhou2023navgpt, lin2024navcot, zhou2024navgpt2}, leveraging generic visual-linguistic representations and alignment~\cite{chen2021history, guhur2021airbert, li2022envedit, li2024panogen, li2023improving, liu2024volumetric,Wang2024GOAT}, and enhancing navigation action planning mechanisms with map construction and historical memorization~\cite{zhao2022target, wang2023gridmm, hong2020recurrent, wang2020active, liu2023bird}. Although there has been considerable progress in this field, the high cost of collecting human-annotated instruction trajectory data to achieve near-human performance is still a key obstacle to training generalizable capable VLN agents. Numerous data augmentation methods have been proposed like training instruction-generator or extracting large-scale datasets from rendered environments in simulators~\cite{hao2020prevalent, zhang2024navhint, ramakrishnan2021hm3d, xia2018gibson, wang2023scalevln, wang2024bootstrapping}, but none of which satisfies the need for large-scale high-quality demonstrations on exploration strategies. To tackle this challenge, we propose self-improving demonstration method for goal-oriented VLN.

\noindent\textbf{Self-Improving Agents} are capable of autonomously enhancing themselves through iterative feedback to achieve continuous improvement without extensive human intervention~\cite{lin1992self, singh2023beyond}. Significant progress has been made in various fields such as mathematical reasoning~\cite{yuan2023scalingrelationshiplearningmathematical, zhao2025sirius}, software engineering~\cite{hu2024self} and robotic manipulation~\cite{bousmalis2023robocat}. The quality of self-generated data can be improved through various techniques, including model's self-feedback~\cite{wu2024meta, yuan2024self}, reward ranked or rejection sampling finetuning~\cite{dong2023raft, zelikman2022star, xiong2025minimalist}. However, in robotics tasks, complicated sampling strategies are impractical due to the intricate evaluation of alignment of instruction-trajectory pairs~\cite{wang2024bootstrapping}, and the key obstacle is the quality of demonstrations rather than correctness (\textit{e.g.}, both exploration trajectories and shortest paths are correct, but the former provides better demonstrations than the latter). In SID, the agent is capable of iteratively improving itself by leveraging its own generated trajectories, eliminating the dependence on extra data generators.

\noindent\textbf{Learning with Demonstrations} to overcome inefficient exploration~\cite{nair2018overcoming} is essential for sparse-reward tasks such as robotic navigation and manipulation. For instance, TCN~\cite{sermanet2018time} offers an approach to learning from video demonstrations for tasks requiring temporal understanding. RoboTube~\cite{xiong2022robotube} showcases the potential to learn household manipulation tasks from human videos demonstrations with simulated environments. More recently, RLDG~\cite{xu2024rldg} distills generalist policies from RL-generated trajectories to reduce human dependency and enhance adaptability. In goal-oriented navigation, Habitat-Web~\cite{ramrakhya2022habitat} leverages human demonstrations to train exploration-aware agents for object navigation. Unlike these approaches, SID bootstraps from the agent's self-demonstrations iteratively, enabling strong exploration policy without external human effort.
\vspace{-2.0pt}
\section{Self-Improving Demonstrations for Goal-oriented VLN} 

Goal-oriented navigation involves planning a navigation agent to reach the target only with its visual / textual description in unseen environments. Existing methods rely on costly human demonstrations to devise feasible exploration strategies towards the targets. To address this data bottleneck, we propose {\model} to scale the training data in a semi-supervised manner. In the following, we detail this method and its training corpus.
\vspace{-2.0pt}
\subsection{Self-Improving Demonstrations on Image Goal Navigation}

\begin{figure*}[t]
  \vspace{-2.0pt}
  \centering
    \includegraphics[width=1.0\textwidth]{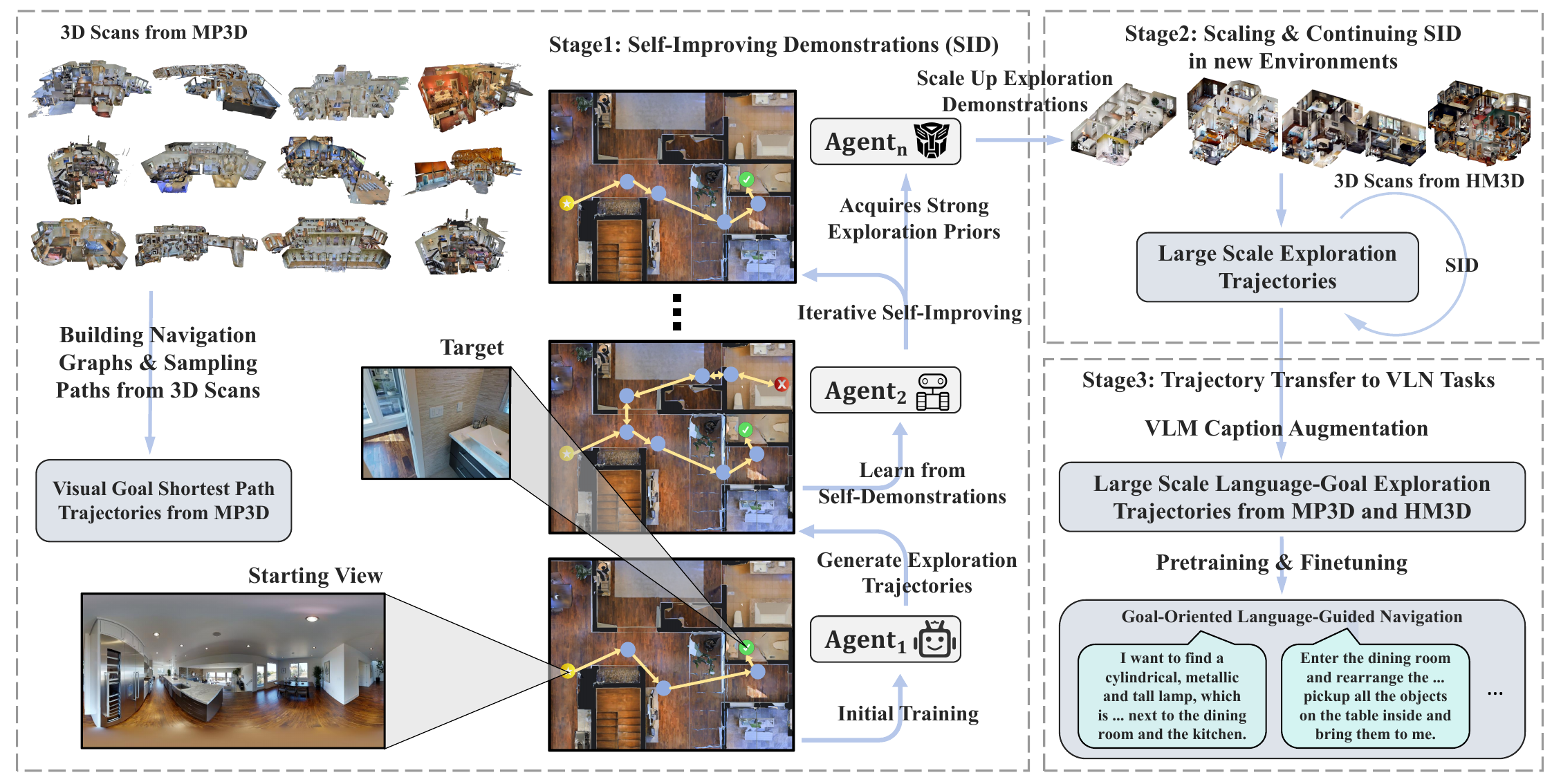}
  \caption{\textbf{Our proposed Self-Improving Demonstrations paradigm for goal-oriented VLN.} We learn an initial navigation agent using trajectories sampled from MP3D, generate new paths using this agent, and reserve the successful exploration ones. These trajectories give demonstrations on the exploration strategies, resulting in a more capable agent. This iterative semi-supervised learning can gradually improve navigation agent's performance ceiling and produce effective exploration trajectories at scale, which can be transferred with caption augmentation for goal-oriented VLN.}
  \label{fig:overall_pipeline}
  \vspace{-12.0pt}
\end{figure*}

We give an approach that continuously improves the quality of goal-oriented navigation training data through an iterative pipeline, as in Figure~\ref{fig:overall_pipeline}. This contains 1) building base training data $D_0$, 2) learning an navigator $\mathcal{N}_{\theta_0}$ with the data, and 3) generating new trajectories from the navigator. This repeats multiple rounds to optimize the navigator's path-planning capabilities as well as goal-oriented navigation data quality. We empirically show it is convergent and consistently improves navigation performance in Section~\ref{sec_exp}.
\vspace{-4.0pt}
\paragraph{Constructing Base Data.} 
We consider navigation in discrete environments. It is formulated as the path planning in graph theory~\cite{anderson2018r2r} and each environment is treated as an undirected graph $\mathcal{G}=\{\mathcal{V},\mathcal{E}\}$, where $\mathcal{V}=\{v_1,v_2,...v_n\}$ denotes $n$ navigable viewpoints and $\mathcal{E}$ denotes connectivity edges. Each viewpoint has a panoramic observation $\mathcal{O}=([\mathbf{I}_i;a_i])_{i=1,..,K}$ of its surroundings and a location coordinate $\mathbf{v}=(x,y,z)$. Note $\mathbf{I}_i$ is the $i$-th view (an image), $a_i$ is the relative angle to face it, and a panorama is composed of $K$ images. We traverse all pairs of nodes in the graph and retain those with the corresponding paths similar to previous human-annotated datasets from MP3D. Specifically, we compute the shortest path between the viewpoints of each pair, and preserve those whose length are between 5 and 7 viewpoints. In this process, we construct a total of over 180,000 trajectories. We associate the panoramic observations to them and get visual-goal-trajectory pairs as raw training data $D_0 = \{(\mathbf{p}_i, \mathbf{g}_i)\}_{i=1,...,n}$, where $\mathbf{p}_i = <\mathbf{v}_i^j>_{j=1,..,l_i}$ is a feasible path reaching the goal $\mathbf{g}_i$, which is the description (\textit{e.g.} an image or a sentence capturing the destination) of the target location $\mathbf{p}_i[-1]$ (the end location of the path $\mathbf{p}_i$). $n$ and $l_i$ denote total sample number and path length, respectively.
Since each viewpoint corresponds to a panorama consisting of 36 images, we ultimately obtain over 6M visual-goal-trajectory pairs.
\vspace{-9pt}
\paragraph{Training Base Navigation Agent.} We utilize the DUET~\cite{chen2022duet}, a widely used navigation model as the agent, which is pretrained from scratch and finetuned on the data $D_0$. Following~\cite{anderson2018r2r}, we treat navigation as a cumulative single action prediction (SAP) process. The agent takes a navigation goal, the current observation, and navigation history as input and outputs the next action decision based on the current node. It is formulated as follows: 
\begin{equation}
    \hat{\mathbf{v}} = \mathcal{N}_{\theta_0}(\mathcal{O}_i^t|\mathbf{g}_i,\mathcal{O}_i^{<t}),
\end{equation}
where $\mathcal{N}$ is the navigation agent parametrized by ${\theta_0}$. $\mathcal{O}_i^t$ is the observation at the $t_\text{th}$ step and $\mathcal{O}_i^{<t}$ is the historical navigation records.

We minimize the imitation loss $\mathcal{L}$ between the agent's predicted actions toward the navigation goal and the ground-truth ones to optimize the agent $\mathcal{N}$ during finetuning, which can be formulated as: 
\begin{equation}
    \theta = \arg\min\limits_{\theta} \left[\sum_{i=1}^n\sum_{t=1}^{l_i} \mathcal{L}(\mathcal{N}_{\theta_0}(\mathcal{O}_i^t|\mathbf{g}_i,\mathcal{O}_i^{<t}),\mathbf{p}_i[t]) \right],
\end{equation}
where $\mathbf{p}_i[t]$ denotes the ground truth location at the step $t$. 

\begin{figure*}[t]
    \vspace{-2.0pt}
    \centering
    \begin{subfigure}[b]{0.32\textwidth}
        \centering
        \includegraphics[width=\textwidth]{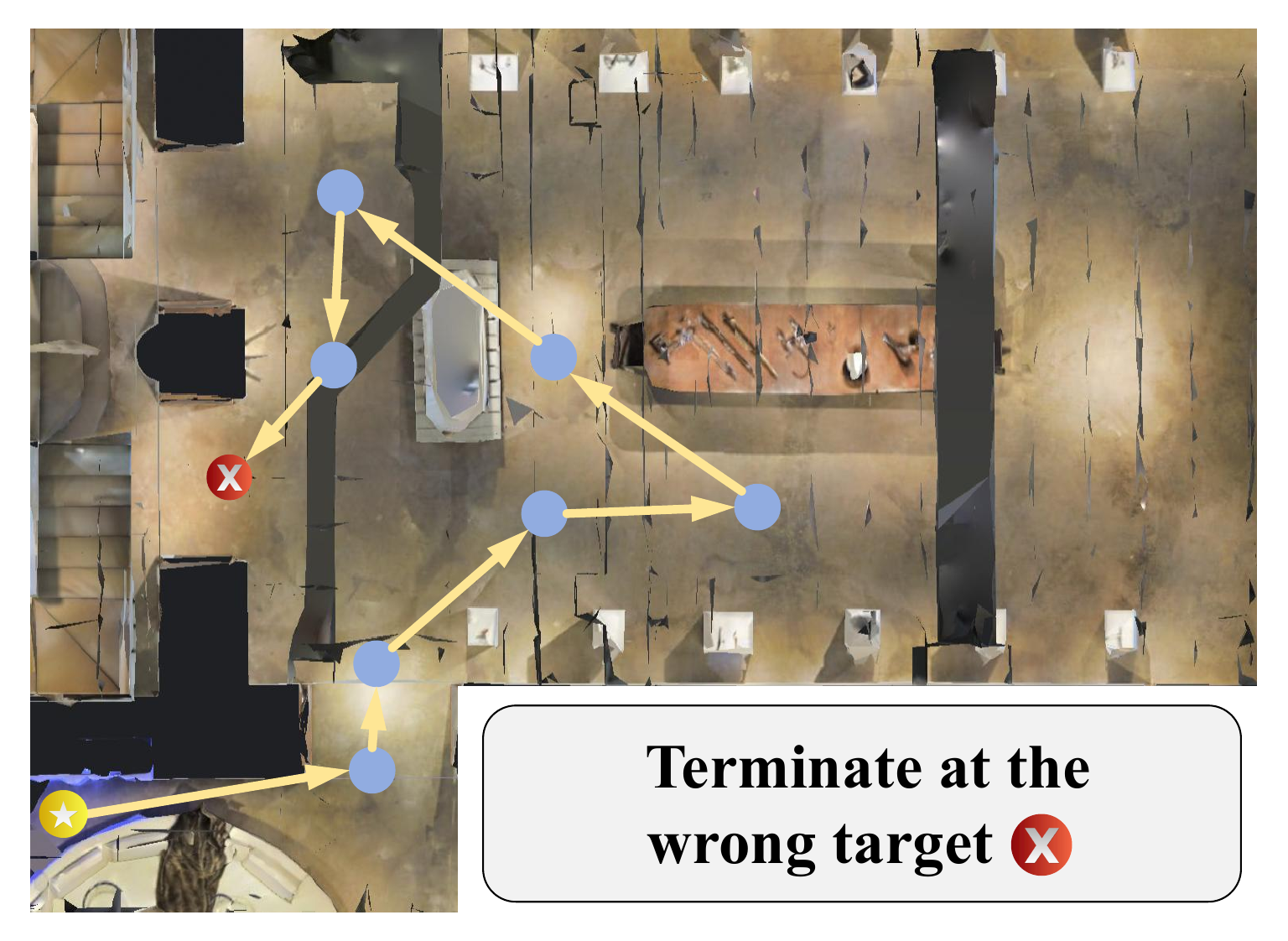}
    \end{subfigure}
    \hfill
    \begin{subfigure}[b]{0.32\textwidth}
        \centering
        \includegraphics[width=\textwidth]{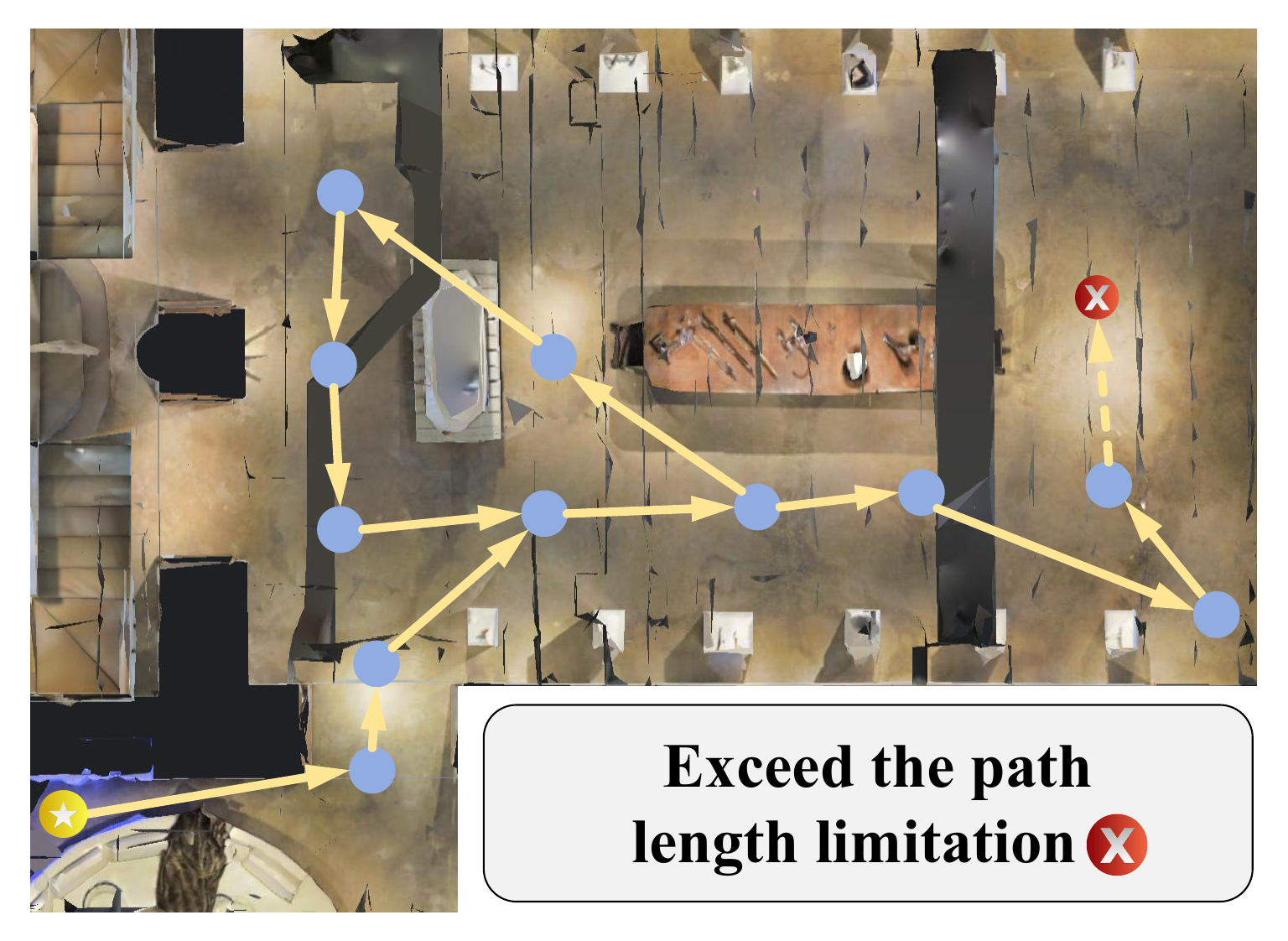}
    \end{subfigure}
    \hfill
    \begin{subfigure}[b]{0.32\textwidth}
        \centering
        \includegraphics[width=\textwidth]{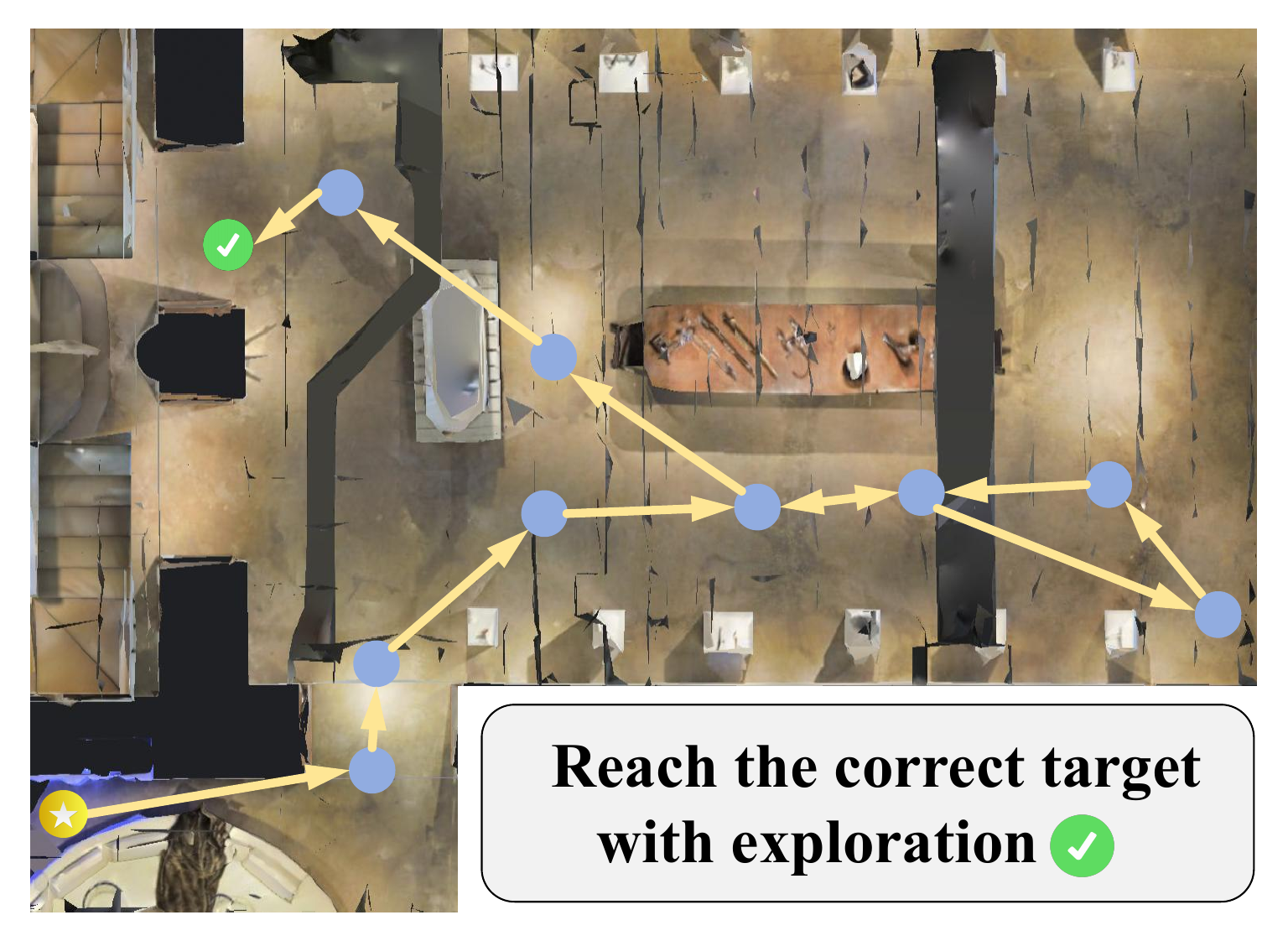}
    \end{subfigure}
    \caption{\textbf{Filtering the trajectories generated by the navigation agent.} The agent may fail in various scenarios, such as terminating at similar but incorrect targets or exceeding the path length limitation. Only the trajectories that successfully reach the correct target with efficient exploration will be retained for subsequent iterations. }
    \label{fig:filter}
   \vspace{-14.0pt}
\end{figure*}

\paragraph{Curating Rollout Data Generation.} Once trained on the shortest-path data, the navigation agent $\mathcal{N}_{\theta_1}$ acquires fundamental exploration capabilities and is utilized to generate new training trajectories. The agent usually needs to flexibly traverse different rooms and distinguish similar but incorrect scenarios before successfully navigating to the goal viewpoint, which encapsulates the agent's exploration of the environment and can serve as the demonstrations to further train the agent. Given the visual goal $g_i$ of the target viewpoint $v_i^t$ and the starting viewpoint $v_i^s$, the agent autonomously generates step-by-step actions based on real-time observation, navigation history and the target. As shown in Figure~\ref{fig:filter}, if the agent reaches the correct target within the path length limitation, the corresponding trajectory will be reserved for the next-iteration training corpus $D_1$.
\vspace{-5.0pt}
\paragraph{Iterative Self-Improving.} We repeat the aforementioned process leveraging the newly generated trajectories as the training set. Specifically, at the iteration $t$, we first utilize the new exploration-training data $D_{t-1}$ for pretraining and finetuning the new stronger navigation agent $\mathcal{N}_{\theta_t}$. Then we use the new agent to refine the $D_0$ to get new training data $D_t$. The reason why we choose to filter the data and train the agent from scratch rather than the previous checkpoint is that we want to eliminate the impact of knowledge in the previous low quality data. Through this self-demonstration iterative mechanism, the agent progressively enhances its exploration capabilities, ultimately yielding a highly capable agent and high-quality dataset for subsequent utilization.

As shown in Figure~\ref{fig:overall_pipeline}, we observe significant trajectory optimization across iterations. Initial shortest-path trajectories tend to contain unstable long steps (\textit{e.g.}, the edge between the third and the fourth viewpoints). In subsequent rounds, the agent learns through exploration, sometimes involving initial misdirection (\textit{e.g.}, entering a similar yet incorrect room) before successfully acquiring the target, to discover more robust, learnable self-generated routes with consistent step lengths. The agent can learn from its own exploration trajectories and reaches the right target directly in the third round. We provide another case of step-by-step trajectory evolution in Section~\ref{sec_4} of the Appendix.

Note that during finetuning, unlike prior methods that rely solely on imitating shortest-path oracle actions via DAgger~\cite{chen2022duet}, we interleave teacher forcing with demonstrations (agent strictly follows the given trajectories) and student forcing with shortest-path oracle actions (agent executes its own predicted actions but is supervised by the oracle). This alternating strategy balances exploration and exploitation, enabling both robust exploration and effective error correction.

\subsection{Scaling and Continuing SID in New Environments}
\label{sec_scale}

Once the navigation agent's performance converges on MP3D environments, we introduce more visual goal trajectories from HM3D~\cite{ramakrishnan2021hm3d} environments and employ the previous agent to generate high-quality large-scale demonstrations on them. The newly generated data are integrated with the previous training data to enable the agent to simultaneously acquire novel exploration capabilities while maintaining its prior navigation knowledge. Considering the computation overhead, we continue the self-improving demonstration process for only one iteration on the whole 800 HM3D environments. Note that to verify the self-improving nature of SID after scaling up, we provide separate experiments of introducing only 60 new HM3D environments beyond the original 60 MP3D scans and continuing SID for two rounds, as detailed in Sec~\ref{sec:img goal main result}.

\subsection{Transfer to Goal-Oriented Vision-and-Language Navigation}

\label{sec_transfer}
As depicted in Figure~\ref{fig:overall_pipeline}, the enhanced navigation agent generates more efficient exploration trajectories, which is then transferred with VLM caption augmentation to align with daily life applications and high-level VLN tasks such as SOON~\cite{zhu2021soon} and REVERIE~\cite{qi2020reverie}. Specifically, we leverage InternVL-26B~\cite{chen2024internvl} to generate semantically rich image descriptions that capture object localization, environmental context and spatial relationships. To maximize data scale while minimizing the redundancy caused by the overlap of different images, we interleavedly choose 18 out of 36 images from the panorama, resulting in 46M high-quality language-goal exploration trajectories. We give a detailed analysis of the choice of caption style in Section~\ref{sec_2} of the Appendix. 

\paragraph{Comparison to Previous VLN datasets.}
\input{tables/data}
Table~\ref{tab:stats} presents detailed statistics of our and previous VLN datasets. As aforementioned, the lack of demonstrations on exploration strategies hinders the advancement of goal-oriented navigation. Existing methods mainly concentrate on generating instructions to tackle data scarcity or improve instruction quality. In contrast, SID focuses on the generation of robust and diverse trajectories. What evolves across different iterations in our approach is not the instructions but the navigation paths, which become increasingly exploratory and efficient through self-improvement. Consequently, the navigation agent can effectively learn exploration strategies from its own trajectories, reduce its error rate, and navigate to the final goal with greater efficiency. In total, our dataset contains over 46M language-goal exploration trajectories with an average of 9.01 viewpoints, which is the first to provide large-scale demonstrations on exploration strategies in goal-oriented VLN.

\section{Experiments}
\label{sec_exp}
\paragraph{Implementation Details.}
We utilize Dinov2~\cite{oquab2023dinov2} and SigLIP~\cite{zhai2023sigmoid} as vision encoders in our experiments following OpenVLA~\cite{kim24openvla}. We utilize LMDeploy to deploy InternVL and annotate a total of 1,213,479 images. We conduct three iterative rounds of SID training on MP3D environments and one iteration of scaling up on 800 HM3D environments. We use the weights of LXMERT~\cite{tan2019lxmert} to initialize DUET, pretrain the agent from scratch for 124k iterations with a batch size of 1024 at a learning rate of $1\times 10^{-5}$ on 8 GPUs, and finetune it on the downstream tasks. The detailed analysis of the computational cost is presented in Table~\ref{tab:compute}, with all experiments conducted on NVIDIA A800 GPUs. Furthermore, to validate the generalization of SID-VLN data across different architectures on continuous environments, we follow VLN-CE~\cite{krantz2020beyond} to transfer the DUET-generated discrete trajectories to continuous ones to train StreamVLN~\cite{wei2025streamvln}, a streaming Video-LLM framework build upon LLaVA-Video~\cite{zhang2024llava} 7B model. We evaluate the trained model on VLN-CE~\cite{krantz2020beyond} benchmark using the Habitat simulator.

\input{tables/compute}

\paragraph{Datasets and Metrics.}
We perform experiments on navigation tasks, including high-level vision-language navigation( REVERIE~\cite{qi2020reverie}, SOON~\cite{zhu2021soon}), Object-Goal Navigation~\cite{savva2019habitat} and VLN-CE~\cite{krantz2020beyond}. The datasets are outlined as follows:

\begin{itemize}[leftmargin=*]
    \item \textbf{REVERIE} consists of a vocabulary of over 1,600 words and 21,702 human-annotated navigational instructions, each describing a target object within MP3D environments~\cite{chang2017matterport3d}. The average length of the collected instructions is 18 words, involving both navigation and referring expression. The shortest path from the agent’s initial location to the target location is between 4 to 7 steps, and the agent needs to find and localize the object in the observation.
    \item \textbf{SOON} includes 3,848 instructions with 1,649-word vocabulary, featuring trajectories in 38 MP3D houses. It also follows the same train/val/test split strategy as REVERIE. The instructions' average length reaches 38.6 words, with trajectories spanning 15-60+ meters and surpassing REVERIE's paths. Due to the increased task difficulty and smaller training dataset size, it is more challenging to achieve high performance on SOON than on REVERIE.
    \item \textbf{Object-Goal Navigation} requires an agent to navigate to the target specified by a general object category. To verify the generalization of SID-VLN to object goal navigation, we follow SAME~\cite{zhou2024same} to transfer the validation split of the Habitat ObjectNav dataset~\cite{savva2019habitat} to evaluate the agent's performance in discrete environments, as detailed in Section~\ref{sec:objnav}.
    \item \textbf{VLN-CE} provides 5.6K trajectories with an average length of 10 meters. It requires agents to navigate through a set of low-level actions (\textit{e.g.} move forward 0.25m, turn-left 15 degrees) rather than teleporting between fixed nodes. 
\end{itemize}

\input{tables/transfer}
\input{tables/scaling}

We evaluate our agent using standard path-fidelity metrics, including Trajectory length (TL), Success Rate (SR), Oracle Success Rate (OSR), Success Rate Weighted by Path Length (SPL)~\cite{anderson2018spl} and Navigation Error (NE). Following~\cite{qi2020reverie}, we adopt Remote Grounding Success (RGS) and RGS weighted by Path Length (RGSPL) to evaluate object grounding. The agent is considered successful only if it stops accurately at the designated target viewpoint in image-goal navigation.

\subsection{Self-Improving Demonstration Running Results}
\paragraph{Consistent Self improvement of navigation agent.} SID generates both novel effective trajectories and enhanced agents from the given environments via self-improving demonstrations. As shown in the left part of Table~\ref{tab:sid}, the SPL on unseen environments improves from 40.24\% to 44.62\%. And the close alignment of OSR and SR indicates the agent possesses a robust capability to identify the target---once it traverses the goal location, it can accurately terminate the episode there. Furthermore, the successfully sampled trajectories by the agent increase while the average length of them continuously decreases from 8.27 to 7.27 concurrently. This highlights the self-bootstrapping nature of SID: the agent iteratively learns better exploration strategies through its own demonstrations. 
\vspace{-5.0pt}
\paragraph{Scalability and suitability to new environments.} When SID converges in MP3D environments, we show it can scale up and continue to self-improve when extended to new environments. (1) Additional navigation trajectories from 800 HM3D environments yields a clear 10\% improvement in SR and SPL in unseen environments as shown in Table~\ref{tab:sid}. (2) Scaling up SID to 60 additional HM3D environments and continuing two rounds of processes shows that SID retains its self-improving nature with increased data in Table~\ref{tab:two_rounds}, demonstrating its effective scalability with larger training environments.
\vspace{-5.0pt}
\paragraph{Room Exploration.}
Given that our navigation tasks are performed in 3D indoor environments consisting of interconnected rooms, evaluating the room exploration and error correction is crucial.
As in Table~\ref{tab:room_exploration}, the agent trained on self-exploration data explores a greater number and variety of rooms than the shortest paths. Besides, the number of rooms explored of the target type has increased, verifying that the agent actively explores similar rooms and corrects its wrong steps to reach the correct destination. The navigation efficiency and success rate continuously improve with the progress of SID, showcasing that the agent effectively learns exploration strategies from its own trajectories, reduces its error rate, and navigates to the target with greater efficiency. We provide a case of step-by-step room exploration and trajectory evolution in Figure~\ref{fig:whole} in the Appendix.
\vspace{-5.0pt}
\paragraph{Transferability to downstream VLN tasks.} 
Table~\ref{tab:sid} reports the performance improvement gained through iterative self-demonstrations on SOON and REVERIE with transferred language-goal navigation training. These results indicate the strong transferability of SID exploration trajectories to goal-oriented VLN tasks.

\subsection{Ablation and Analysis}

\paragraph{Supervision Strategies.}
\label{sec:supervision}
We compare different supervision strategies in the teacher forcing in Table~\ref{tab:supervision}. (1) Self-demonstrations benefits finetuning across different data scales and goal modals, revealing the significance of learning from self-exploration strategies instead of greedily taking the shortest paths. (2) Even finetuned with the same shortest-path demonstrations, the image-goal navigation agent pretrained on self-demonstrations (Table~\ref{tab:supervision}, line 1) still performs significantly better than the one pretrained on shortest-path demonstrations (Table~\ref{tab:sid}, line 1), which underscores that self-demonstrations benefit both pretraining and finetuning.

\vspace{-5.0pt}
\paragraph{Demonstration Sources.} We show that the navigation agent gains increase from self-demonstrations rather than random exploration. We sample random paths for goal-oriented navigation data. The number of viewpoints of these trajectories ranges from 7 to 11 and the average number is 8.9, similar to the distribution of the agent's exploration ones. Table~\ref{tab:random_ablate} shows that the agent trained on the randomly sampled trajectories can only reach the equivalent performance with the shortest path but far behind the exploration paths of the agent itself, which strongly claims that the agent learns effectively from its own exploration rather than arbitrary trajectories, validating SID for goal-oriented navigation.
\vspace{-5.0pt}
\paragraph{Effectiveness of SID beyond Simple Data Scaling.} To verify the effect of the agent's demonstrations in new environment rather than simply scaling up, we conduct another comparison by merely adding shortest-path trajectories in HM3D environments to train a new agent. The results in Table~\ref{tab:scale_ablate} shows that incorporating shortest path data modestly improves performance but is distinctly surpassed compared to leveraging the agent's exploration trajectories, which demonstrates that the primary benefit arises from the SID paradigm rather than simply introducing more data in new environments.
\vspace{-5.0pt}
\paragraph{Limitation.} SID's exploration can fail in some intricate scenarios. If the agent continuously explores the environment, it will exceed the maximum navigation steps limitation and be forced to stop. Furthermore, SID agent is pretrained on robotic navigation assumptions, which may need additional adaptation for practical applications. And the successful transfer of SID to VLN tasks relies on captions generated by VLMs, which may be incomplete or inaccurate due to hallucinations. We present the detailed analysis in Section~\ref{sec_5} of the Appendix.

\label{sec:img goal main result}
\input{tables/supervision_sid}
\input{tables/ablate}

\subsection{Comparisons on Downstream VLN tasks}
\label{sec:main}
\paragraph{SOON and REVERIE.} As shown in Table~\ref{tab:soon}, SID reaches new state-of-the-art results across all navigation metrics on the challenging SOON task, exceeding AutoVLN with fewer training 3D environments. Notably, on the SOON test-unseen leaderboard, SID outperforms all previous methods and achieves the highest success rate.
On the REVERIE task, Table~\ref{tab:rvr} shows that SID also achieves new state-of-the-art performance, which underscores the significance of efficient demonstrations on exploration strategies instead of simple instruction augmentation for challenging goal-oriented VLN. Furthermore, the trajectory length of SID proves that our enhancement stems from genuinely more efficient exploration routes rather than merely allowing successful yet longer episodes.

\input{tables/soon}

\input{tables/obj}
\paragraph{Object-Goal Navigation.}
\label{sec:objnav}
 As Table~\ref{tab:objnav} shows, the navigation agent trained on SID-VLN data notably achieves a 76\% success rate on the transferred objectnav validation unseen set, surpassing DUET by a large relative margin of 8\%. This indicates that the exploration capabilities learned by SID from self-demonstrations can effectively generalize to object-goal navigation. The experimental details are presented in Section~\ref{sec_3} of the Appendix.

\input{tables/vlnce}

\paragraph{VLN-CE.} The results in Table~\ref{tab:vlnce} indicates that the model augmented with SID-VLN data learns to actively search the environment to locate the goal, resulting in substantial improvements in NE, OSR, and SR, accompanied by a slight decrease in SPL (due to more exploration). This demonstrates the strong transferability of SID's exploration priors to different model architecture on VLN-CE.

\section{Conclusion}
In this paper, we introduce SID for learning goal-oriented navigation with Self-Improving Demonstrations at scale. SID is a scalable and transferrable learning paradigm that enables agents to iteratively enhance their navigation capabilities through self-generated, exploration-rich trajectories. Leveraging agent exploration as supervision, SID mitigates the reliance on costly human annotations for large-scale demonstrations. Our experiments indicate a consistent improvement of SID in the classical image-goal navigation task. Additionally, we show that SID trajectories can be seamlessly transformed into vision-and-language navigation with VLM caption augmentation and elevate the new state-of-the-art performance across diverse goal-oriented VLN tasks, demonstrating great potential for unified goal-oriented navigation. And SID-VLN data also showcases distinct transferability across different models and on object-goal navigation and VLN-CE. In general, SID sets a new standard for scalable goal-oriented navigation learning and opens promising directions for self-improving embodied learning.

\clearpage
%
%
\bibliographystyle{splncs04}
\bibliography{main}

\setcounter{section}{0}
\newpage
\vspace{-5.0pt}
\section*{Appendix}

We first describe the implementation details of our experiments in Section~\ref{sec_1}, including pretraining objectives and details of Self-Improving Demonstrations experiments. In Section~\ref{sec_2}, we provide detailed experiments about transferring the learned SID exploration priors to vision-language navigation tasks and the effects of caption styles. Section~\ref{sec_3} presents the experimental details of downstream navigation tasks. Section~\ref{sec_4} visualizes our navigation trajectories towards the target. In Section~\ref{sec_5}, we discuss some potential limitations of our SID method.

\section{Implementation Details}

\label{sec_1}

\subsection{Navigation Agent}

We employ the DUal-scale Graph Transformer (DUET)~\cite{chen2022duet}, a well-established and widely used navigation model as the agent. The core innovation of DUET is its dual-scale reasoning mechanism fusing information from both coarse and fine-grained encoders. The coarse-grained global encoder operating on the global topological map leverages a graph-aware self-attention mechanism to effectively incorporate the intrinsic map structure, enabling robust reasoning over extended distances and complex spatial layouts. Meanwhile, the fine-grained local encoder processes detailed local visual information from the current viewpoint (including multi-view panoramic imagery and object features) for precise language grounding and local action choices. This combination allows DUET to effectively balance nuanced instruction comprehension with strategic long-range planning.

\subsection{Pretraining Objectives}

 We mainly employ two proxy tasks, MLM and SAP, to pretrain the agent. Here we describe these two proxy tasks in detail. The inputs for these tasks are goal-trajectory pairs as $D = \{(\mathbf{p}_i, \mathbf{g}_i)\}_{i=1,...,n}$, where $\mathbf{p}_i = <\mathbf{v}_i^j>_{j=1,..,l_i}$ is the demonstration trajectories reaching the goal $\mathbf{g}_i$ and $\mathbf{g}_i$ is the description (\textit{e.g.}, an image or a sentence capturing the destination) of the target location $\mathbf{p}_i[-1]$ (the end location of the path $\mathbf{p}_i$). $n$ and $l_i$ denote total sample number and path length, respectively. Among them, SAP is utilized throughout the multi round self-improving demonstrations, while MLM is adopted only during transferring and we sample one task for each iteration with equal probability.

\paragraph{Masked Language Modeling (MLM)} involves predicting masked words based on the provided textual context and the full navigation trajectory. In this task, the description $\mathbf{g}_i$ of the target location is equivalent to instruction $\mathcal{W} = \{w_1,w_2,\dots,w_n\}$. A special \verb|[mask]| token is used to randomly mask out 15\% of the tokens in $\mathcal{W}$. We predict the masked word distribution $p (w_i|\hat{\mathcal{W}}, \mathcal{P})=f_{\text{MLM}}(x_i)$ through a standard two-layer fully-connected network, where $\hat{\mathcal{W}}$ is the masked instruction,  $x_i$ is the output embedding of the masked word $w_i$ and $\mathcal{P}$ is the corresponding trajectory. The objective is to minimize the negative log-likelihood of predicting the original words: $\mathcal{L}_{\text{MLM}} = - \mathrm{log}\ p (w_i|\hat{\mathcal{W}}, \mathcal{P})$. 

\paragraph{Single Action Prediction (SAP)} aims to predict the next action based on the goal description and the given navigation history path. Following~\cite{chen2022duet}, we predict the probability for each candidate action in the action space via a two-layer fully-connected network. The optimization objective is to minimize the negative log probability of the ground-truth target view action $\mathcal{L}_{\mathrm{SAP}} = -\mathrm{log}\ p_t(a_t | \mathcal{W}, \mathcal{P}_{<t})$. 

Specifically, we refine the sampling strategy for selecting viewpoints during training to enhance both path adherence and error correction. The original probability distribution in DUET is 20\% for the goal viewpoint (to learn stopping), 40\% for other on-path viewpoints (to learn path progression), and 40\% for random off-path viewpoints (to learn error correction). However, considering that on-path points constitute only a very small fraction of the total navigable viewpoints within an environment, random sampling of off-path points provides sparse and often uninformative negative examples for effective error correction. Furthermore, our finetuned models demonstrate a high Oracle Success Rate (OSR) that closely aligns with their Success Rate (SR), which indicates that the agent is highly proficient at making the decision to stop if it reaches the correct final viewpoint. 

Based on these key observations, we introduce a revised and more balanced sampling strategy. We allocate 75\% of our sampling probability to uniformly selecting points along the self-demonstration trajectories, on which we want the agent to focus more.  The remaining 25\% of the sampling probability is dedicated to selecting challenging negative examples specifically from the candidate navigable viewpoints. This targeted negative sampling strategy is informed by an analysis of agent failures on 2000 unseen evaluation trajectories. We observed a distinct error distribution across trajectory steps (\textit{e.g.}, [927, 295, 253, 189, 214, 98, 24] errors at steps 1 to N, respectively). Therefore, our revised strategy prioritizes sampling difficult negatives (\textit{i.e.}, off-path candidate viewpoints where the agent is likely to err), providing more effective supervision for learning robust error correction and subsequently improving overall navigation policy.

\begin{figure*}[!t]
  \centering
  \includegraphics[width=0.98\textwidth]{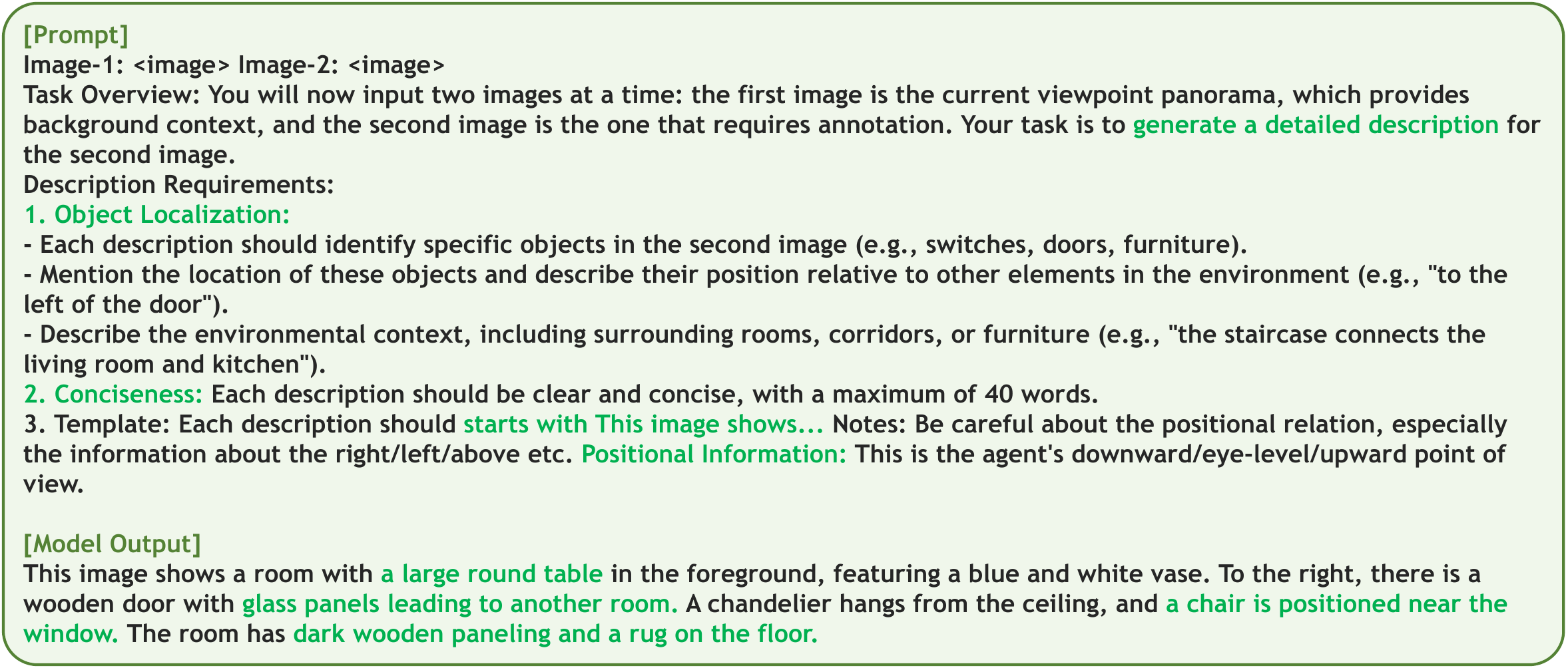}
  \caption{Prompt and Model Output of the Detail-style Captions.}
  \label{fig:detail}
  \vspace{1.0em}

  \begin{minipage}{0.48\textwidth}
    \centering
    \includegraphics[width=\textwidth]{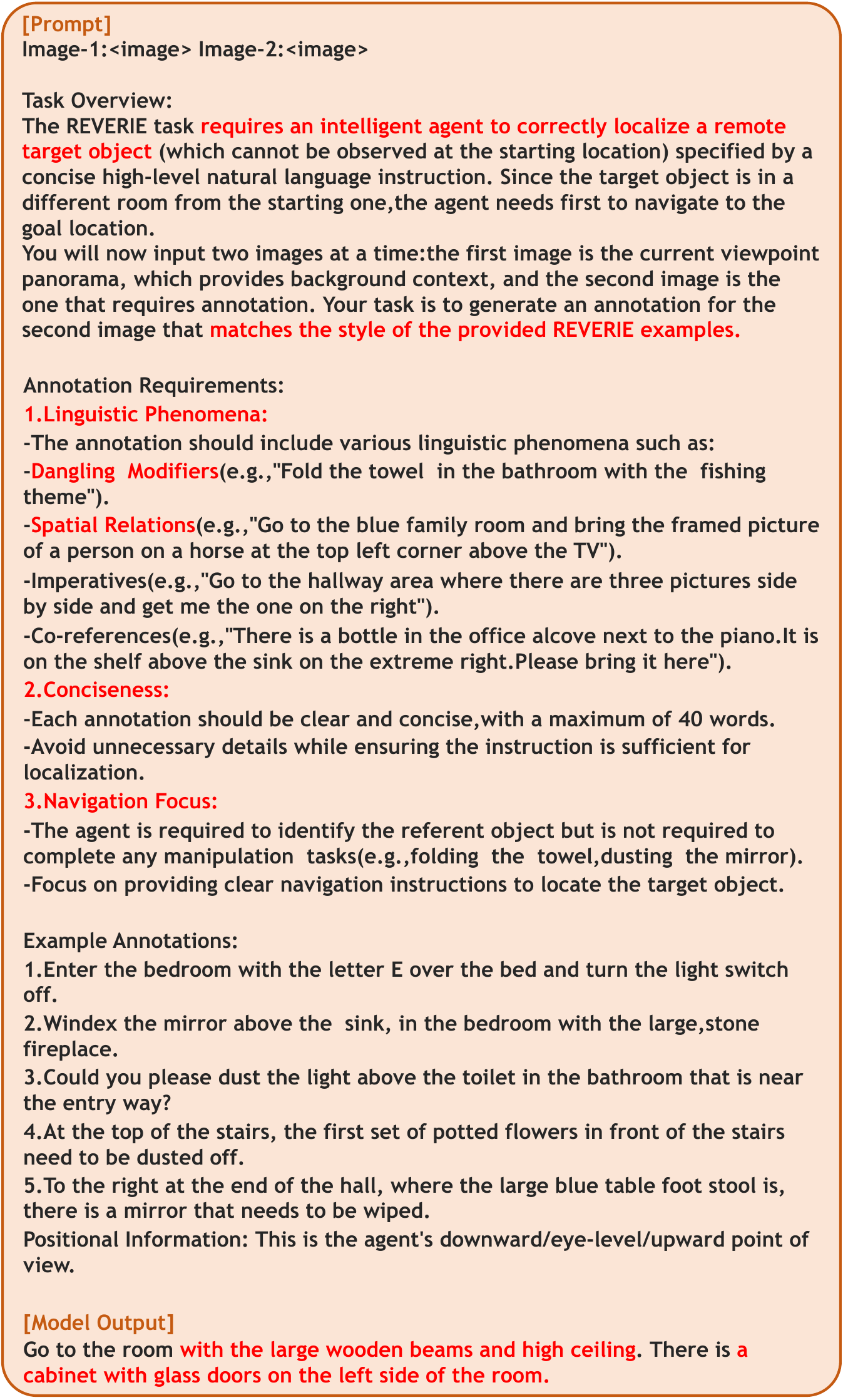}
  \caption{Prompt and Model Output of the REVERIE-style Captions.}
  \label{fig:rvr}
  \end{minipage}
  \hfill
  \begin{minipage}{0.48\textwidth}
    \centering
    \includegraphics[width=\textwidth]{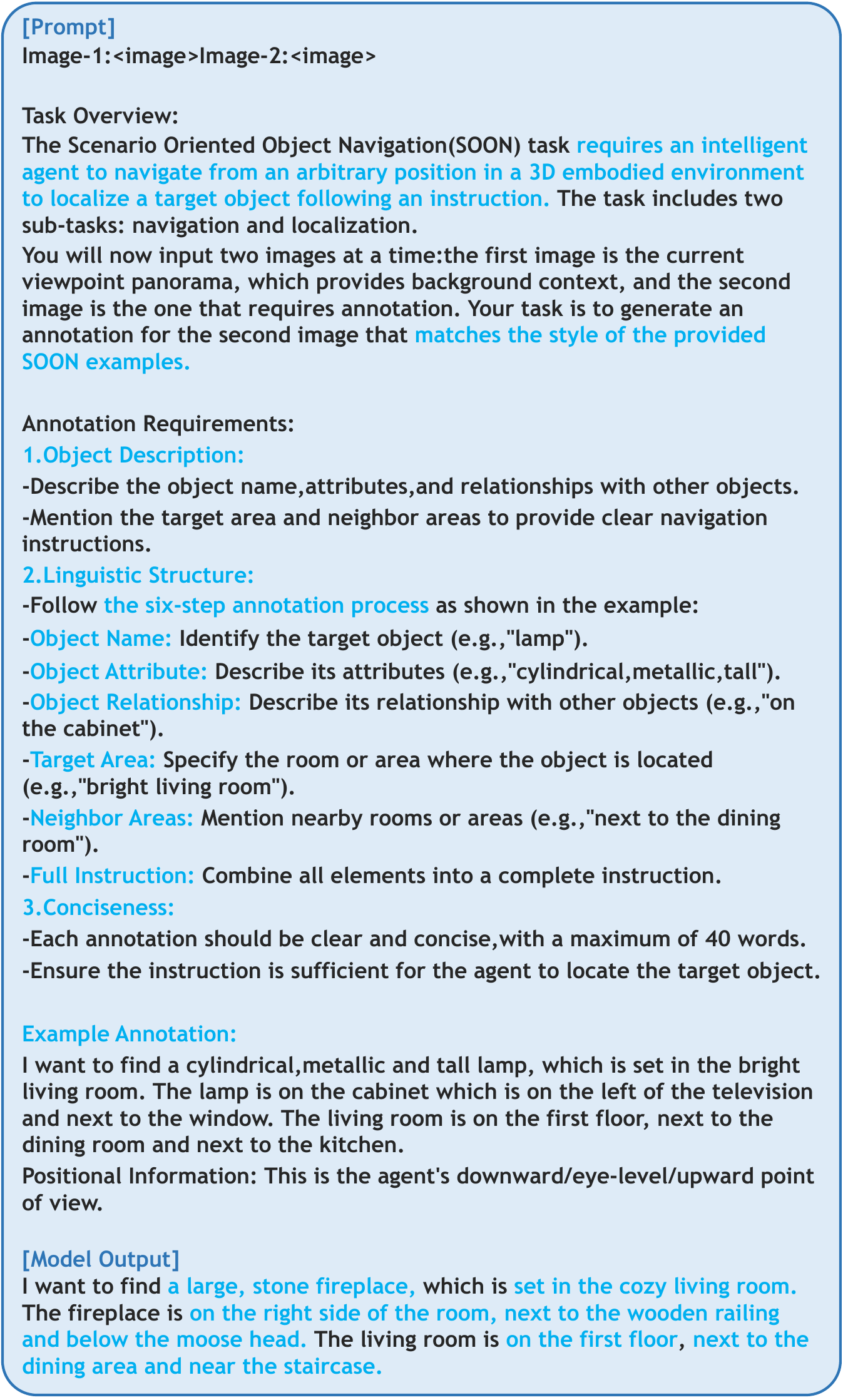}
  \caption{Prompt and Model Output of the SOON-style Captions.}
  \label{fig:soon}
  \end{minipage}  
  \label{fig:example_layout}
\end{figure*}

\subsection{Details of experiments of Self-Improving Demonstration}

Goal-oriented vision-language navigation contains trajectories that lead to target objects specified by high-level instructions. Following their distribution of the data, for every visible object at a target viewpoint, we sample paths with a length between 5 and 7 viewpoints that end at the target, resulting in 181,246 paths from MP3D environments. Since each viewpoint corresponds to a panorama consisting of 36 images, we ultimately obtain 6,524,856 visual-goal navigation trajectories. Then we utilize the weights of LXMERT~\cite{tan2019lxmert} to initialize DUET and pretrain this model with our synthetic visual-goal datasets.

We utilize Dinov2-Base~\cite{oquab2023dinov2} and SigLIP-Base~\cite{zhai2023sigmoid} as the visual encoders to extract the image features. For the first round, we pretrain DUET for 200k iterations with a batch size of 128 and a learning rate of $5\times 10^{-5}$ on MP3D environments. We compare the model checkpoints and pick the one with the highest finetuning performance. Then we finetune DUET for 300k iterations, with batch size 64 and learning rate $5\times 10^{-5}$ on 8 NVIDIA A800 GPUs.

We repeat the aforementioned process for three rounds to get a strong navigation agent, which is then adopted to generate new trajectories for the visual goals in the original MP3D and scale up in new 800 HM3D environments. We follow the same sampling rule as in MP3D environments, resulting in 3,220,504 paths and 115,938,144  visual-goal-trajectory pairs from HM3D. Then we use the navigation agent trained through self-improving demonstrations to generate new paths and obtain totally 60M visual-goal trajectories from 860 MP3D and HM3D environments. Based on the new training corpus, we continue the above SID process for another round and get the final self-demonstration trajectories.

\section{Transferring to vision-language navigation tasks}
\label{sec_2}

To transfer the generated trajectories with exploratory demonstration to vision-language navigation tasks, we leverage VLM to generate captions for the corresponding images of the target. Specifically, to maximize the overall data scale and minimize the semantic redundancy caused by the spatial overlap of different images, we choose 18 out of 36 images from the panorama in an interleaved way. We first employ InternVL-26B to generate three styles of caption on MP3D environments to explore the most suitable style for caption augmentation:

\begin{itemize}[leftmargin=*]
    \item Detail-style: This prompt generates detailed descriptions with object localization (\textit{e.g.}, "to the left of the door"), environmental context (\textit{e.g.}, "the staircase connects the living room") and positional relation (\textit{e.g.}, "the left/right/above"), providing the navigator fine-grained visual information, as in Figure ~\ref{fig:detail}.

    \item REVERIE-style: This prompt generates REVERIE-like annotations, using linguistic phenomena such as dangling modifiers and spatial relations and focuses on concise navigation instructions for Remote Embodied Visual Referring Expression in Real Indoor Environments~\cite{qi2020reverie}, as depicted in Figure ~\ref{fig:rvr}.

    \item SOON-style: This prompt produces annotations with object names, attributes (\textit{e.g.}, "cylindrical") and relationships as well as target and neighbor areas(\textit{e.g.}, "next to the dining room"), strictly following the setting of Scenario Oriented Object Navigation (SOON)~\cite{zhu2021soon}, with the template shown in Figure ~\ref{fig:soon}.
\end{itemize}

After getting three styles of caption, we transfer the training data by replacing the images with corresponding captions but keeping the trajectories created by the navigation agent, resulting in 3M caption-trajectories training corpus. Although the REVERIE-Style and SOON-Style captions are more similar to the instructions in the downstream VLN tasks, the Detail-Style captions lead to the best zero-shot and finetune performance on both REVERIE and SOON as shown in Table ~\ref{tab:ft}. The agent pretrained on detail-style caption-trajectory pairs surpasses the other ones after being finetuned on the downstream tasks and the baseline model (pretrained on initial, non-caption data) on SOON. It also reaches comparable performance on REVERIE. And the competitive zero-shot performance also demonstrates the navigation agent's capabilities of visual-language-action alignment gained from Detail-style captions pretraining.

With the empirical results above, we finally generate large-scale detailed-style caption on HM3D environments and leverage the transferred detail-caption training data to pretrain DUET for 124k iterations with a batch size of 1024 and a learning rate of $1\times10^{-5}$ with SAP and MLM. The pretrained agent is then utilized to finetune on SOON, REVERIE and Object-Goal Navigation. 

\begin{table*}[t]
\centering
\begin{minipage}{0.49\textwidth}
    \centering
    \setlength{\tabcolsep}{3pt}
    \renewcommand{\arraystretch}{1.27}
    \caption{\textbf{Zero-shot performance} in unseen environments on REVERIE with different styles of pretraining data.}
    \label{tab:zero}
    \resizebox{\textwidth}{!}{
    \begin{tabular}{l|ccccc}
    \Xhline{1.0pt}
     \multicolumn{1}{c|}{\multirow{2}{*}{Styles}} & \multicolumn{5}{c}{REVERIE} \\
     \cline{2-6} & \multicolumn{1}{c}{OSR$\uparrow$} & \multicolumn{1}{c}{SR$\uparrow$} &
    \multicolumn{1}{c}{SPL$\uparrow$} & \multicolumn{1}{c}{RGS$\uparrow$} &
    \multicolumn{1}{c}{RGSPL$\uparrow$} \\
    \Xhline{1.0pt}  
     REVERIE & 38.57 & 24.03 & 15.43 & 4.23 & 2.79 \\ 
     SOON & 41.86 & 27.26 & 14.67 & 4.74 & 2.47 \\
    \rowcolor{blue!10} \textbf{Detail} & \textbf{52.88} & \textbf{29.59} & \textbf{16.94} & \textbf{5.54} & \textbf{3.59} \\
     \Xhline{1.0pt}
\end{tabular}
    }
\end{minipage}
\hfill
\begin{minipage}{0.49\textwidth}
    \centering
    \setlength{\tabcolsep}{3pt} 
    \renewcommand{\arraystretch}{1.27} 
    \caption{\textbf{Finetune performance} in unseen environments on REVERIE and SOON with different styles of pretraining data.}
    \label{tab:ft}
    \resizebox{\textwidth}{!}{
    \begin{tabular}{l|ccc|ccc}
    \Xhline{1.0pt}
     \multicolumn{1}{c|}{\multirow{2}{*}{Styles}} & \multicolumn{3}{c|}{REVERIE} &\multicolumn{3}{c}{SOON} \\
     \cline{2-7} & \multicolumn{1}{c}{OSR$\uparrow$} & \multicolumn{1}{c}{SR$\uparrow$} &
    \multicolumn{1}{c|}{SPL$\uparrow$} & \multicolumn{1}{c}{OSR$\uparrow$} &
    \multicolumn{1}{c}{SR$\uparrow$} &
    \multicolumn{1}{c}{SPL$\uparrow$}\\
    \Xhline{1.0pt}  
     DUET & 53.59 & 50.58 & 38.27 & 54.57 & 39.23 & 26.99 \\ 
     \Xhline{0.5pt}
     SOON & 52.88 & 47.63 & 34.69 & 48.82 & 39.53 & 26.29 \\
     REVERIE & 50.36 & 44.33 & 32.45 & 48.67 & 37.91 & 28.16 \\ 
    \rowcolor{blue!10} \textbf{Detail} & \textbf{55.10} & \textbf{50.87} & \textbf{36.56} & \textbf{54.49} & \textbf{46.02} & \textbf{30.58} \\
     \Xhline{1.0pt}
\end{tabular}
    }
\end{minipage}
\end{table*}

\section{Experimental Details of the downstream navigation tasks.}
\label{sec_3}

Here we provide experimental details on downstream tasks, with the results presented in the main paper. For REVERIE, we finetune the DUET model with a batch size of 16, learning rate of $3\times10^{-5}$ and ml\_weight of $0.1$ with our SID data augmentation for 20k iterations. On SOON, we use a batch size of 8, learning rate of $3\times10^{-5}$ and ml\_weight of $0.2$ without augmentation for 5k iterations.

For Object-Goal Navigation task, we follow the methodology in SAME~\cite{zhou2024same} to adapt the Habitat-MP3D dataset~\cite{savva2019habitat} by discretizing its continuous environment into a topological connectivity graph $\mathcal{G}$. Trajectories from Habitat-Web are then matched to the nearest nodes on $\mathcal{G}$ based on minimum Euclidean distance, with sequentially repeated nodes merged, resulting in a total of 58,803 trajectories with an average of 20 steps. We similarly transfer data from the MP3D validation split to evaluate model performance in these discrete environments. After pretraining the DUET with the transferred data, we finetune it and the SID agent on this Object-Goal Navigation task, as presented in the main paper.
\label{sec_objnav}

To validate the cross-architecture generalization of our SID-VLN data in continuous environments, we follow VLN-CE~\cite{krantz2020beyond} to transfer the generated discrete trajectories to continuous ones to train StreamVLN~\cite{wei2025streamvln}, which is a streaming Video-LLM framework employing a hybrid slow-fast context modeling strategy to support multi-modal reasoning over interleaved vision, language and action inputs. Unlike discrete settings, VLN-CE requires agents to execute low-level actions (\textit{e.g.}, turn left or turn right by 15 degrees, move forward by 25 centimeters, or stop) within the Habitat simulator using raw RGB-D observations, rather than teleporting between discrete nodes. Following ~\cite{wei2025streamvln}, the baseline in our paper is trained on the navigation trajectories from R2R-CE~\cite{anderson2018r2r}, R2R-EnvDrop~\cite{tan2019envdrop} as well as ScaleVLN~\cite{wang2023scalevln}. And we augment this navigation agent with our transferred data. By integrating with SID-VLN exploration priors, the agent learns to actively search environments, leading to substantial improvements in success rate on the validation unseen split, as depicted in the main paper.

\section{Visualization of the trajectories}
\label{sec_4}
\begin{figure*}
  \vspace{-5.0pt}
  \centering
  \begin{minipage}[c]{0.45\textwidth}
    \centering
    \includegraphics[width=\textwidth, height=17cm]{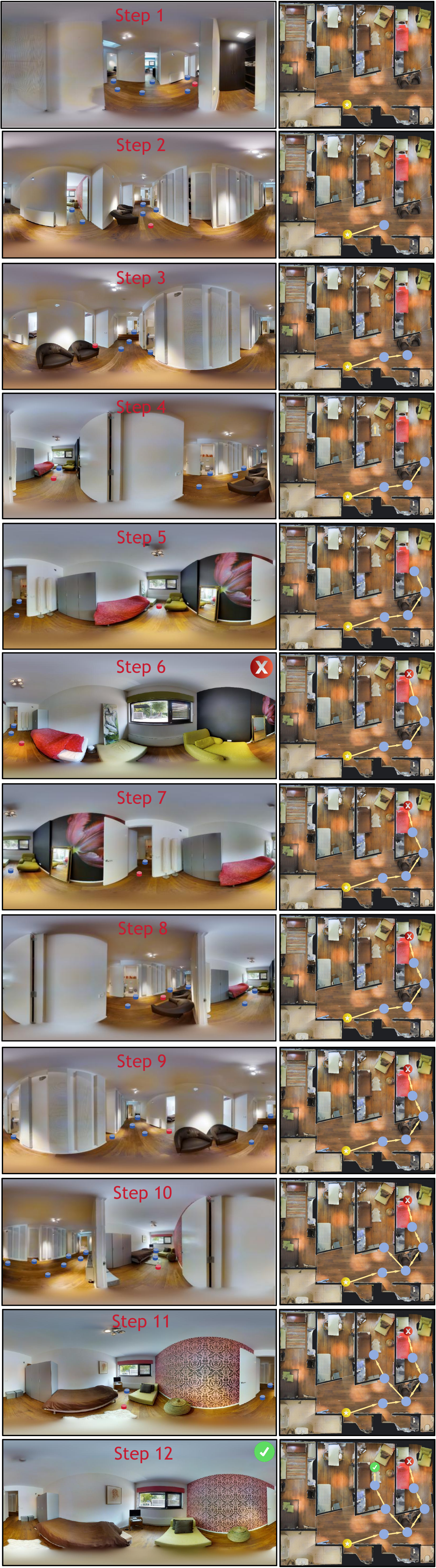}
    \label{fig:left}
  \end{minipage}
  \hfill
  \begin{minipage}[c]{0.44\textwidth}
    \centering
    \includegraphics[width=\textwidth, height=9cm]{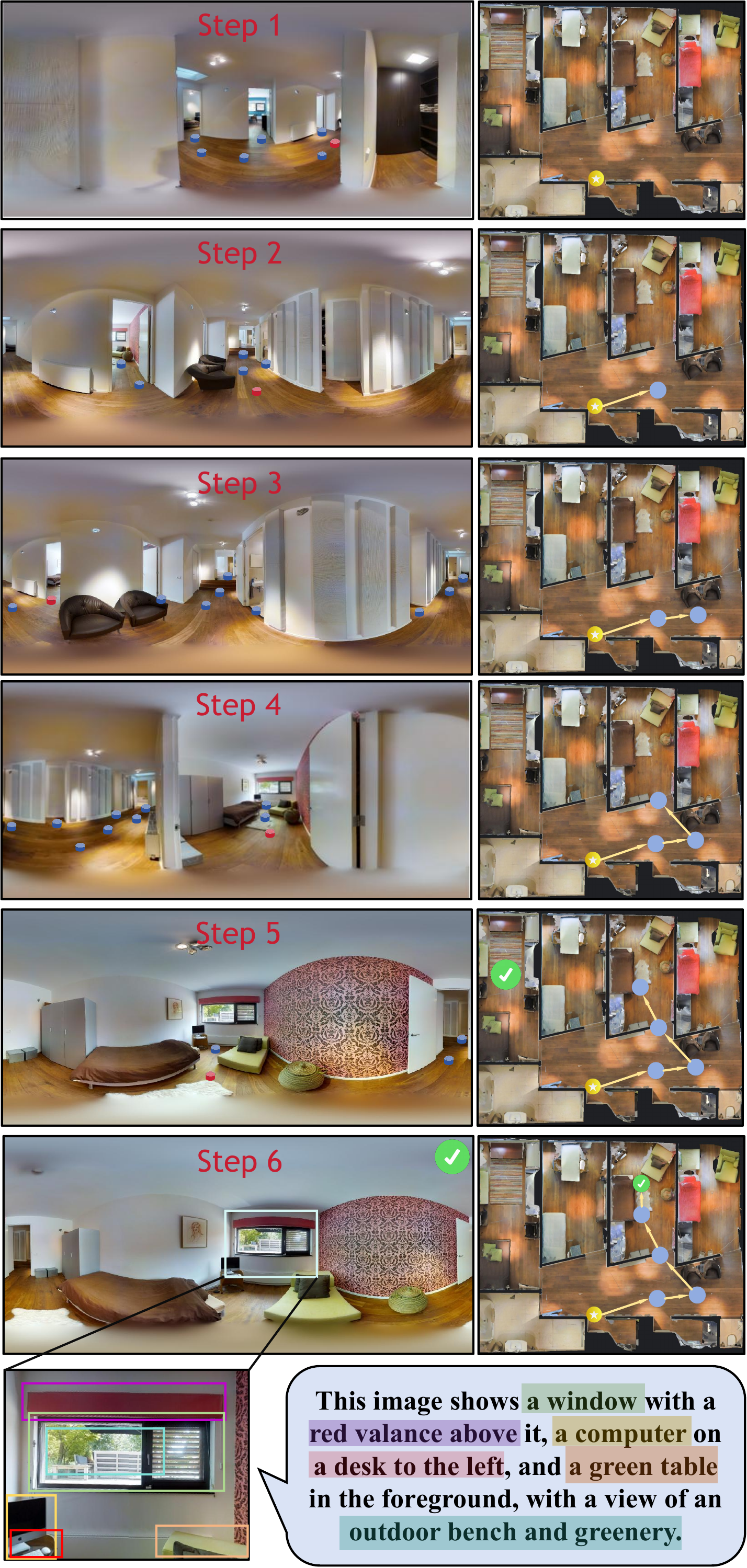}
    \label{fig:final}
    
    \includegraphics[width=\textwidth, height=7.5cm]{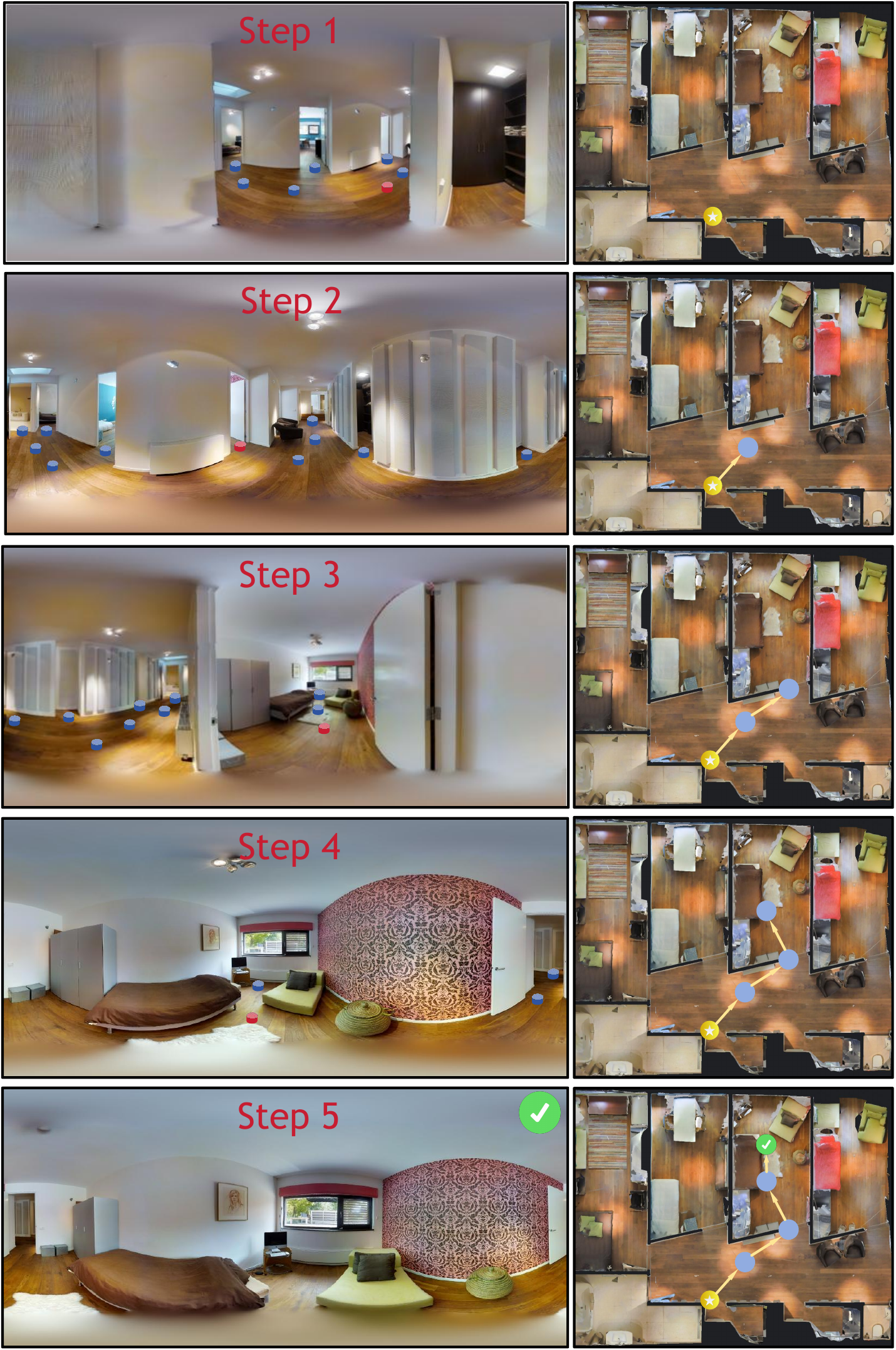}
    \label{fig:short}
  \end{minipage}
  \vspace{-5.0pt}
  \caption{Visualization of the navigation agent's exploration trajectory and the predicted actions at each step. Right-Bottom: The initial shortest-path trajectory towards the navigation target. Left: The navigation agent's exploration and error-correction trajectory across different rooms. Right-Up: The navigation agent's final trajectory and the corresponding image and caption of the target.}
  \label{fig:whole}
\end{figure*}

Figure ~\ref{fig:whole} visualizes the agent’s exploration trajectories and the corresponding predicted actions at each step during multi-round self-improving. At the first round, the navigation agent learns primarily from the sampled shortest-path trajectory. But the capability gained cannot help the agent to distinguish between two adjacent highly-similar rooms. So in the next round, the agent actively explores the two rooms and finds out the first room lacks critical elements (\textit{e.g.}, the red valance and the computer on the desk to the left). Then it returns and explores the other room, correctly reaching the intended target location. The navigation agent learns from its own demonstration on the exploration strategies and picks the right room directly in the third round, which reveals the agent's robust capability of fine-grained visual grounding and trajectory optimization as well as the advantage of SID in sophisticated goal-oriented navigation tasks.

\section{Limitations}
\label{sec_5}
SID's exploration can fail in complex areas with numerous choices at critical navigational viewpoints. If the agent continuously explores the environment, it will exceed the maximum navigation steps limitation and then be forced to stop, hindering success in intricate exploration scenarios. It's important to note that previous methods also frequently struggle with these challenging cases. In our future work, we plan to enhance SID's capabilities of error correction and efficient exploration to further boost success rate in difficult scenarios.

Moreover, we acknowledge the following limitations and directions for future work. Firstly, there is a potential embodiment gap as the agent is pretrained on specific robotic navigation assumptions, which may not directly translate to real-world robot deployment. Further challenges include the inherent sim-to-real gap, necessitating additional adaptation for practical robotic applications. Furthermore, SID's successful transfer to VLN tasks hinges largely on textual descriptions generated by VLMs, which may produce incomplete or inaccurate captions due to potential hallucinations. And evaluating caption quality in the navigation domain is particularly challenging as it requires assessing the alignment between vision, language, and the actions an agent might take. Developing robust mechanisms for further verifying and refining these generated captions in the embodied tasks could enhance robustness and fully combine the vision-language alignment capabilities of VLMs to bootstrap unified goal-oriented navigation.

\end{document}

%% file: tables/data.tex
\begin{wraptable}{r}{0.56\textwidth}
    \vspace{-35pt}
    \caption{\resizebox{0.84\linewidth}{!}{\textbf{Statistics of different VLN training data.}}}
    \label{tab:stats}
    \centering
    \renewcommand{\arraystretch}{1.1}
    \setlength{\tabcolsep}{1mm}
    \begin{adjustbox}{width=\linewidth}
    \begin{tabular}{l|c|c|ccc} \Xhline{1.0pt}
        Dataset & Generated & Demonstration & \#Env. &  \#Data \\ \hline
        \multicolumn{5}{@{}l}{\darkGreen{\textit{Fine-grained VLN}}} \\
        R2R~\cite{anderson2018r2r} & \ding{55} & - & 61 & 14,039 \\
        RxR-en~\cite{anderson2020rxr} & \ding{55} & - & 60 & 26,464 \\
        Marky~\cite{wang2022less} & \ding{51} & - & 60 & 333,777 \\        
        \hline
        \multicolumn{5}{@{}l}{\darkGreen{\textit{Goal-oriented VLN}}} \\
        REVERIE~\cite{qi2020reverie} & \ding{55} & \ding{55} & 60 & 10,466 \\
        SOON~\cite{zhu2021soon} & \ding{55} & \ding{55} & 34 & 27,800 \\
        AutoVLN~\cite{chen2022hm3dlearning} & \ding{51} & \ding{55} & 900 & 217,703 \\
        ScaleVLN-RVR~\cite{wang2023scalevln} & \ding{51} & \ding{55} & 1289 & 831,318 \\
        NavRAG~\cite{wang2025navrag}  & \ding{51} & \ding{55} & 861 & 2,115,019 \\
        \hline
        \rowcolor{blue!10} SID-46M (Ours) & \ding{51} & \ding{51} & 860 & 46,534,355 \\
        \Xhline{1.0pt}
    \end{tabular}
    \end{adjustbox}
    \vspace{-20pt}
\end{wraptable}

%% file: tables/compute.tex
\begin{table*}[t]
\centering
\setlength{\tabcolsep}{4pt}
\renewcommand{\arraystretch}{1.2}
\caption{\textbf{Quantitative analysis of the computation overhead.} Caption, Feature, Trajectory and Pretraining denotes VLM-based Caption Generation, Visual Feature Extraction, Trajectory Sampling in all iterations of SID training and the final Language-Goal Pretraining on SID-46M, respectively.}
\label{tab:compute}
\resizebox{1.0\textwidth}{!}{
\begin{tabular}{l|c|c|c|cccc|c}
    \Xhline{1.0pt}
    \multicolumn{1}{c|}{\multirow{2}{*}{Stages}} &\multicolumn{1}{c|}{\multirow{2}{*}{Caption}} &\multicolumn{1}{c|}{\multirow{2}{*}{Feature}} &\multicolumn{1}{c|}{\multirow{2}{*}{Trajectory}} & \multicolumn{4}{c|}{SID Training} & \multicolumn{1}{c}{\multirow{2}{*}{Pretraining}} \\
    \cline{5-8}
    \multicolumn{1}{c|}{} & \multicolumn{1}{c|}{} &\multicolumn{1}{c|}{} & \multicolumn{1}{c|}{} & 
    \multicolumn{1}{c}{Round1} & \multicolumn{1}{c}{Round2}& \multicolumn{1}{c}{Round3} & \multicolumn{1}{c|}{Scale} & \multicolumn{1}{c}{} \\
    \Xhline{1.0pt}
    GPU Hours & 460 & 190 & 1500 & 140 & 180 & 260 & 1200 & 1500 \\
    \Xhline{1.0pt}
    \end{tabular}
}
\vspace{-8pt}
\end{table*}

%% file: tables/transfer.tex
\begin{table*}[t]
\centering
\setlength{\tabcolsep}{6pt} 
\renewcommand{\arraystretch}{1.1}
\caption{
\textbf{Detailed analysis of multi-round SID Learning.} Left: Performance on Image Goal Navigation validation unseen split and statistics of the training trajectories data. Right: Performance in unseen environments on REVERIE and SOON. \# VP denotes the number of viewpoints.
}
\label{tab:sid}
\resizebox{\textwidth}{!}{
\begin{tabular}{c|ccc|cc|cc|cc}
\Xhline{1.0pt}
\multicolumn{1}{c|}{\multirow{3}{*}{Round}} 
& \multicolumn{5}{c|}{Image Goal Navigation Training} & \multicolumn{4}{c}{Transferred VLN Performance} \\
\cline{2-10}
& \multicolumn{3}{c|}{Val Unseen} & \multicolumn{2}{c|}{Traj Statistics} & \multicolumn{2}{c|}{REVERIE} & \multicolumn{2}{c}{SOON}\\
\cline{2-10}
\multicolumn{1}{c|}{} & OSR$\uparrow$ & SR$\uparrow$ & SPL$\uparrow$ & Traj Num & Avg \#VP & SR$\uparrow$ & SPL$\uparrow$ & SR$\uparrow$ & SPL$\uparrow$ \\
\Xhline{1.0pt}
\multicolumn{7}{@{}l}{\darkGreen{\textit{SID in MP3D environments}}} \\
1 & 61.85 & 61.45 & 40.24 & 5,082,528 & 8.27 & 50.87 & 36.56 & 45.02 & 30.58 \\
2 & 65.50 & 65.35 & 43.31 & 5,698,944 & 7.31 & - & - & - & - \\
\rowcolor{darkgray!10}
3 & 65.55 & 65.45 & 44.62 & 5,873,474 & 7.27 & 51.92 & 39.80 & 46.46 & 34.51 \\
\multicolumn{7}{@{}l}{\darkGreen{\textit{Scaling up with 800 HM3D  environments}}} \\
\rowcolor{blue!10} \textbf{4}
 & \textbf{75.00} & \textbf{75.00} & \textbf{54.67} & 93,069,068 & 9.01 & \textbf{59.39} & \textbf{45.95} & \textbf{50.88} & \textbf{38.42} \\
\Xhline{1.0pt}
\end{tabular}
}
\end{table*}

%% file: tables/scaling.tex
\begin{table*}[t]
\centering
\begin{minipage}{0.55\textwidth} 
    \centering
    \setlength{\tabcolsep}{3pt} 
    \renewcommand{\arraystretch}{1.27} 
    \caption{\textbf{Performance on Image Goal Navigation when continuing SID with 60 additional HM3D environments for two iterations.}}
    \label{tab:two_rounds}
    \resizebox{\textwidth}{!}{
    \begin{tabular}{c|ccc|ccc}
    \Xhline{1.0pt}
     \multicolumn{1}{c|}{\multirow{2}{*}{Tot Env}} & \multicolumn{3}{c|}{Validation Seen} &\multicolumn{3}{c}{Validation Unseen} \\
     \cline{2-7} & \multicolumn{1}{c}{OSR$\uparrow$} & \multicolumn{1}{c}{SR$\uparrow$} &
    \multicolumn{1}{c|}{SPL$\uparrow$} & \multicolumn{1}{c}{OSR$\uparrow$} &
    \multicolumn{1}{c}{SR$\uparrow$} &
    \multicolumn{1}{c}{SPL$\uparrow$}\\
    \Xhline{1.0pt}  
    60 & 87.40 & 86.60 & 79.18 & 65.55 & 65.45 & 44.62 \\
    120 (Round 1) & 86.80 & 86.80 & 78.38 & 70.00 & 69.90 & 48.58 \\
     \rowcolor{blue!10} \textbf{120 (Round 2)} & \textbf{89.20} & \textbf{88.60} & \textbf{79.24} & \textbf{70.80} & \textbf{70.75} & \textbf{48.60} \\ 
     \Xhline{1.0pt}
\end{tabular}
    }
\end{minipage}
\hfill
\begin{minipage}{0.43\textwidth} 
    \centering
    \setlength{\tabcolsep}{4pt} 
    \renewcommand{\arraystretch}{1.27} 
    \caption{\textbf{Statistics of room exploration of the trajectories on the image goal navigation validation unseen split.}}
    \label{tab:room_exploration}
    \resizebox{\textwidth}{!}{
    \begin{tabular}{c|ccc}
    \Xhline{1.0pt}
     \multicolumn{1}{c|}{\multirow{2}{*}{Trajectory Source}} & \multicolumn{3}{c}{Room Statistics} \\
     \cline{2-4} & \multicolumn{1}{c}{Num} & \multicolumn{1}{c}{Type} & \multicolumn{1}{c}{Target}\\
    \Xhline{1.0pt}  
     Original Shortest Paths & 2.77 & 2.62 & 1.09 \\
     Agent trained on MP3D & 4.21 & 3.05 & 1.38 \\
     \rowcolor{blue!10} Agent scaled on HM3D & 3.89 & 2.94 & 1.32 \\
     \Xhline{1.0pt}
\end{tabular}
    }
\end{minipage}
\vspace{-5pt}
\end{table*}

%% file: tables/supervision_sid.tex
\begin{table*}[t]
\centering
\setlength{\tabcolsep}{4pt} 
\renewcommand{\arraystretch}{1.2}
\caption{\textbf{Comparison of different strategies of supervision in teacher forcing.} Left: Performance on image goal navigation with MP3D data finetuning. Right: Performance on REVERIE validation unseen split with large-scale SID data augmentation.}
\label{tab:supervision}
\resizebox{\textwidth}{!}{
\begin{tabular}{c|ccc|ccc|cccc|cc}
\Xhline{1.0pt}
\multicolumn{1}{c|}{\multirow{3}{*}{Strategies}} 
& \multicolumn{6}{c|}{Image Goal Navigation Finetuning} & \multicolumn{6}{c}{Augmented REVERIE Finetuning} \\
\cline{2-13}
& \multicolumn{3}{c|}{Validation Seen} & \multicolumn{3}{c|}{Validation Unseen} & \multicolumn{4}{c|}{Navigation} & \multicolumn{2}{c}{Grounding}\\
\cline{2-13}
\multicolumn{1}{c|}{} & OSR$\uparrow$ & SR$\uparrow$ & SPL$\uparrow$ & OSR$\uparrow$ & SR$\uparrow$ & SPL$\uparrow$ & TL$\downarrow$ & OSR$\uparrow$ & SR$\uparrow$ & SPL$\uparrow$& RGS$\uparrow$ & RGSPL$\uparrow$ \\
\Xhline{1.0pt}
Shortest & 81.80 & 80.40 & 68.32 & 65.05 & 64.90 & \textbf{44.63} & 20.95 & \textbf{63.62} & 58.53 & 44.04 & 37.18 & 27.93 \\
\rowcolor{blue!10}  \textbf{Explored} & \textbf{83.00} & \textbf{82.40} & \textbf{71.40} & \textbf{65.50} & \textbf{65.35} & 43.31 & \textbf{20.78} & 63.45 & \textbf{59.39} & \textbf{45.95} & \textbf{38.54} & \textbf{29.69} \\
\Xhline{1.0pt}
\end{tabular}
}
\end{table*}

%% file: tables/ablate.tex
\begin{table*}[!t]
\centering
\begin{minipage}{0.49\textwidth} 
    \centering
    \setlength{\tabcolsep}{3pt}
    \renewcommand{\arraystretch}{1.27} 
    \caption{\textbf{Comparison on VLN validation unseen splits with different demonstration sources.}}
    \label{tab:random_ablate}
    \resizebox{\textwidth}{!}{
  \begin{tabular}{c|ccc|ccc}
    \Xhline{1.0pt}
    \multicolumn{1}{c|}{\multirow{2}{*}{Sources}} & \multicolumn{3}{c|}{REVERIE} & \multicolumn{3}{c}{SOON}\\
    \cline{2-7}  & \multicolumn{1}{c}{OSR$\uparrow$} & \multicolumn{1}{c}{SR$\uparrow$} & \multicolumn{1}{c|}{SPL$\uparrow$}& \multicolumn{1}{c}{OSR$\uparrow$} & \multicolumn{1}{c}{SR$\uparrow$} & \multicolumn{1}{c}{SPL$\uparrow$}\\ 
    \Xhline{1.0pt}
    Shortest & 55.10 & 50.87 & 36.56 & 54.49 & 46.02 & 30.58 \\
    Random & 54.22 & 50.87 & 37.04 & 52.36 & 42.18 & 32.48 \\ 
    \Xhline{0.5pt}
    \rowcolor{blue!10} \textbf{Agent} & \textbf{56.09} & \textbf{51.92} & \textbf{39.80} & \textbf{55.28} & \textbf{46.46} & \textbf{34.51} \\
    \Xhline{1.0pt}
  \end{tabular}
    }
\end{minipage}
\hfill
\begin{minipage}{0.49\textwidth} 
    \centering
    \setlength{\tabcolsep}{3pt} 
    \renewcommand{\arraystretch}{1.27} 
    \caption{\textbf{Comparison on VLN validation unseen splits with different scaling strategies.}}
    \label{tab:scale_ablate}
    \resizebox{\textwidth}{!}{
  \begin{tabular}{c|ccc|ccc}
    \Xhline{1.0pt}
    \multicolumn{1}{c|}{\multirow{2}{*}{Strategies}} & \multicolumn{3}{c|}{REVERIE} & \multicolumn{3}{c}{SOON}\\
    \cline{2-7}  & \multicolumn{1}{c}{OSR$\uparrow$} & \multicolumn{1}{c}{SR$\uparrow$} & \multicolumn{1}{c|}{SPL$\uparrow$}& \multicolumn{1}{c}{OSR$\uparrow$} & \multicolumn{1}{c}{SR$\uparrow$} & \multicolumn{1}{c}{SPL$\uparrow$}\\ 
    \Xhline{1.0pt}
    \textit{w/o} & 56.09 & 51.92 & 39.80 & 55.28 & 46.46 & 34.51 \\
    Shortest & 60.97 & 55.16 & 42.58 & 56.43 & 48.35 & 37.65\\ 
    \Xhline{0.5pt}
    \rowcolor{blue!10} \textbf{Explored} & \textbf{63.45} & \textbf{59.39} & \textbf{45.95} & \textbf{60.32} & \textbf{50.88} & \textbf{38.42} \\
    \Xhline{1.0pt}
  \end{tabular}
    }
\end{minipage}
\vspace{-5.0pt}
\end{table*}

%% file: tables/soon.tex
\begin{table*}[t]
\centering
\setlength{\tabcolsep}{2pt} 
\renewcommand{\arraystretch}{1}
\vspace{-5pt}
\caption{
\textbf{Comparison with the state-of-the-art methods on SOON datasets.}
}
\label{tab:soon}
\resizebox{0.87\textwidth}{!}{
\begin{tabular}{l|ccccc|ccccc}
\Xhline{1.0pt}
\multicolumn{1}{c|}{\multirow{2}{*}{Methods}} 
& \multicolumn{5}{c|}{Validation Unseen}
& \multicolumn{5}{c}{Test Unseen}
\\
\cline{2-11}
\multicolumn{1}{c|}{} 
 & \multicolumn{1}{c}{TL$\downarrow$} & \multicolumn{1}{c}{OSR$\uparrow$} & \multicolumn{1}{c}{SR$\uparrow$} & \multicolumn{1}{c}{SPL$\uparrow$} & \multicolumn{1}{c|}{RGSPL$\uparrow$} & \multicolumn{1}{c}{TL$\downarrow$} & \multicolumn{1}{c}{OSR$\uparrow$} & \multicolumn{1}{c}{SR$\uparrow$} & \multicolumn{1}{c}{SPL$\uparrow$} & \multicolumn{1}{c}{RGSPL$\uparrow$} \\
\Xhline{1.0pt}
Human
& - & - & - & - & - & - & 91.4 & 90.4 & 59.2 & 51.1 \\
\Xhline{0.5pt}
DUET~\cite{chen2022duet} & 36.2 & 50.9 & 36.3 & 22.6 & 3.8 & 41.8 & 43.0 & 33.4 & 21.4 & 4.2 \\
NaviLLM~\cite{zheng2024navillm} & - & - & 38.3 & 29.2 & - & - & - & 35.0 & 26.3 & - \\
KERM~\cite{li2023kerm} & 35.8 & 51.6 & 38.1 & 23.2 & 4.0 & - & - & - & - & - \\  
GridMM~\cite{wang2023gridmm} & 38.9 & 53.4 & 37.5 & 24.8 & 3.9 & 46.2 & 48.0 & 36.3 & 21.3 & 4.2 \\
MBA~\cite{zhang2024seeing} & 37.2 & - & 42.0 & 29.6 & 6.1 & 36.5 & - & 38.8 & 26.2 & 6.3 \\
GOAT~\cite{Wang2024GOAT} & - & 54.7 & 40.4 & 28.1 & 5.1 & - & 50.6 & 40.5 & 25.2 & 6.1 \\
AutoVLN~\cite{chen2022hm3dlearning} & - & 53.2 & 41.0 & 30.7 & 4.1 & - & 48.7 & 40.4 & 27.8 & 5.1 \\
Meta-Explore~\cite{hwang2023meta} & - & 52.7 & 44.7 & 34.8 & \textbf{8.9} & - & 48.7 & 39.1 & 25.8 & 4.0 \\  
\Xhline{0.5pt}
\rowcolor{blue!10} \textbf{SID (Ours)}
 & \textbf{33.3} & \textbf{60.3} & \textbf{50.9} & \textbf{38.4} & 6.9 & \textbf{36.4} & \textbf{54.0} & \textbf{47.7} & \textbf{35.2} & \textbf{8.3} \\
\Xhline{1.0pt}
\end{tabular}
}

\vspace{0.4cm}

\caption{
\textbf{Comparison with the state-of-the-art methods on REVERIE datasets.}
$^*$ denotes utilizing SigLIP and Dinov2 as vision encoders for fair comparison.
}
\label{tab:rvr}
\resizebox{\textwidth}{!}{
\begin{tabular}{l|cccccc|cccccc}
    \Xhline{1.0pt}
    \multicolumn{1}{c|}{\multirow{2}{*}{Methods}} & \multicolumn{6}{c|}{Validation Unseen} & \multicolumn{6}{c}{Test Unseen} \\
    \cline{2-13}
    \multicolumn{1}{c|}{} & \multicolumn{1}{c}{TL$\downarrow$} & \multicolumn{1}{c}{OSR$\uparrow$} & \multicolumn{1}{c}{SR$\uparrow$} & \multicolumn{1}{c}{SPL$\uparrow$} & \multicolumn{1}{c}{RGS$\uparrow$} & \multicolumn{1}{c|}{RGSPL$\uparrow$} & 
    \multicolumn{1}{c}{TL$\downarrow$} & \multicolumn{1}{c}{OSR$\uparrow$} & \multicolumn{1}{c}{SR$\uparrow$} & \multicolumn{1}{c}{SPL$\uparrow$} & \multicolumn{1}{c}{RGS$\uparrow$} & \multicolumn{1}{c}{RGSPL$\uparrow$} \\
    \Xhline{1.0pt}
    Human
    & - & - & - & - & - & - & 21.2 & 86.8 & 81.5 & 53.7 & 77.8 & - \\
    \Xhline{0.5pt}
    DUET~\cite{chen2022duet}
    & 22.1 & 51.1 & 47.0 & 33.7 & 32.1 & 23.0 & 21.3 & 56.9 & 52.5 & 36.1 & 31.9 & 22.1 \\
    NaviLLM~\cite{zheng2024navillm}
    & - & 53.7 & 44.6 & 36.6 & - & - & - & 56.2 & 43.5 & 34.5 & - & -  \\
    BEVBert~\cite{an2023bevbert}
    & - & 56.4 & 51.8 & 36.4 & 34.7 & 24.4 & - & 57.3 & 52.8 & 36.4 & 32.1 & 22.1 \\
    BSG~\cite{liu2023bird} 
    & 24.7 & 58.1 & 52.1 & 35.6 & 35.4 & 24.2 
    & 22.9 & 62.8 & 56.5 & 38.7 & 33.2 & 22.3 \\
    VER~\cite{liu2024volumetric}
    & 23.0 & 61.1 & 56.0 & 39.7 & 33.7 & 23.7 & 24.7 & 62.2 & 56.8 & 38.8 & 33.9 & 23.2 \\
    GOAT~\cite{Wang2024GOAT}
    & - & - & 53.4 & 36.7 & 38.4 & 26.1 & - & - & 57.7 & 40.5 & \textbf{38.3} & 26.7 \\
    \Xhline{0.5pt}
    \multicolumn{7}{@{}l}{\darkGreen{\textit{Paradigm with the same 800 environments from HM3D}}} \\
    AutoVLN~\cite{chen2022hm3dlearning}
    & - & 62.1 & 55.9 & 40.9 & 36.6 & 26.8 & - & 62.3 & 55.2 & 38.9 & 32.2 & 22.7 \\
    ScaleVLN~\cite{wang2023scalevln}
    & - & 63.9 & 57.0 & 41.8 & 35.8 & 26.1 & - & 62.7 & 56.1 & 39.5 & 32.5 & 22.8 \\
    NavRAG~\cite{wang2025navrag}
    & - & \textbf{70.7} & 57.3 & 42.0 & - & - & - & - & - & - & - & - \\
    ScaleVLN$^*$~\cite{wang2023scalevln}
    & - & 64.8 & 58.4 & 43.5 & 37.7 & 28.1 & - & - & - & - & - & - \\
    \Xhline{1.0pt}
    \rowcolor{blue!10} \textbf{SID (Ours)}
    & \textbf{20.8} & 63.5 & \textbf{59.4} & \textbf{46.0} & \textbf{38.5} & \textbf{29.7}
    & \textbf{20.2} & \textbf{65.5} & \textbf{60.4} & \textbf{45.9} & 36.6 & \textbf{27.7} \\
    \Xhline{1.0pt}
    \end{tabular}
}

\vspace{-9pt}
\end{table*}

%% file: tables/obj.tex
\begin{wraptable}{r}{0.32\textwidth}
    \vspace{-32pt}
    \caption{\textbf{Comparison on the Transferred ObjectNav MP3D Validation split.}}
    \label{tab:objnav}
    \centering
    \renewcommand{\arraystretch}{1.1}
    \setlength{\tabcolsep}{1.5mm}
    \begin{adjustbox}{width=\linewidth}
    \begin{tabular}{c|cccc}
        \Xhline{1.0pt}
         \multicolumn{1}{c|}{\multirow{2}{*}{Agent}} & \multicolumn{4}{c}{ObjectNav-MP3D(Val)} \\
         \cline{2-5} & \multicolumn{1}{c}{TL$\downarrow$} & \multicolumn{1}{c}{NE$\downarrow$} 
        & \multicolumn{1}{c}{SR$\uparrow$} & \multicolumn{1}{c}{SPL$\uparrow$}\\
        \Xhline{1.0pt}  
         DUET & \textbf{22.17} & 3.67 & 68 & 29 \\
         \rowcolor{blue!10} \textbf{SID} & 24.12 & \textbf{2.74} & \textbf{76} & \textbf{37} \\ 
         \Xhline{1.0pt}
    \end{tabular}
    \end{adjustbox}
    \vspace{-15pt}
\end{wraptable}

%% file: tables/vlnce.tex
\begin{table*}[t]
\centering
\renewcommand{\arraystretch}{1} 
\setlength{\tabcolsep}{2pt}  
\caption{\textbf{Comparison with state-of-the-art methods on the VLN-CE benchmark.} $^*$ indicates methods using the waypoint predictor from~\cite{hong2022bridging}. $\dagger$ denotes using extra transferred SID-VLN training data beyond the VLN-CE datasets.}
\label{tab:vlnce}
\resizebox{0.75\textwidth}{!}{
\begin{tabular}{l|cccc|cccc}
\Xhline{1.0pt}
\multirow{2}{*}{Method} & \multicolumn{4}{c|}{Observation Encoder} & \multicolumn{4}{c}{R2R Val-Unseen} \\ 
\cline{2-9}
& Pano. & Odo. & Depth & S.RGB & NE$\downarrow$ & OS$\uparrow$ & SR$\uparrow$ & SPL$\uparrow$ \\ 
\Xhline{1.0pt}
HPN+DN$^*$~\cite{krantz2021waypoint} & $\checkmark$ & $\checkmark$ & $\checkmark$ &  & 6.31  & 40.0  & 36.0  & 34.0  \\
CMA$^*$~\cite{hong2022bridging}      & $\checkmark$ & $\checkmark$ & $\checkmark$ &  & 6.20  & 52.0  & 41.0  & 36.0  \\
VLN BERT$^*$~\cite{hong2022bridging} & $\checkmark$ & $\checkmark$ & $\checkmark$ &  & 5.74  & 53.0  & 44.0  & 39.0  \\
GridMM$^*$~\cite{wang2023gridmm}    & $\checkmark$ & $\checkmark$ & $\checkmark$ &  & 5.11  & 61.0  & 49.0  & 41.0  \\
ETPNav$^*$~\cite{an2023etpnav}      & $\checkmark$ & $\checkmark$ & $\checkmark$ &  & 4.71  & 65.0  & 57.0  & 49.0 \\ 
ScaleVLN$^{*}$~\cite{wang2023scaling} & $\checkmark$ & $\checkmark$ & $\checkmark$ &  & 4.80  & --    & 55.0  & 51.0 \\
InstructNav~\cite{long2024instructnav} & - & - & - & - & 6.89 & --   & 31.0 & 24.0 \\
R2R-CMTP~\cite{chen2021topological}  & $\checkmark$ & $\checkmark$ & $\checkmark$ &  & 7.90  & 38.0  & 26.4  & 22.7  \\
LAW~\cite{raychaudhuri2021law}       &  & $\checkmark$ & $\checkmark$ & $\checkmark$ & 6.83  & 44.0  & 35.0  & 31.0  \\
ETPNav + FF~\cite{wang2024sim}  &  & $\checkmark$ & $\checkmark$ & $\checkmark$ & 5.95  & 55.8  & 44.9  & 30.4  \\
Seq2Seq~\cite{Krantz2020BeyondTN}    &  &  & $\checkmark$ & $\checkmark$ & 7.77  & 37.0  & 25.0  & 22.0  \\
CMA~\cite{Krantz2020BeyondTN}        &  &  & $\checkmark$ & $\checkmark$ & 7.37  & 40.0  & 32.0  & 30.0  \\
NaVid~\cite{zhang2024navid}   &  &  &  & $\checkmark$ & 5.47  & 49.1  & 37.4  & 35.9  \\
MapNav~\cite{zhang2025mapnav}    &  &  &  & $\checkmark$ & \textbf{4.93}  & 53.0  & 39.7  & 37.2  \\
NaVILA~\cite{cheng2024navila}   &  &  &  & $\checkmark$ & 5.37  & 57.6  & 49.7  & 45.5  \\
StreamVLN   &  &  &  & $\checkmark$ & 5.43  & 62.5  & 52.8 & \textbf{47.2}  \\
\Xhline{1.0pt}
\rowcolor{blue!10}\textbf{StreamVLN}${\dagger}$   &  &  &  & $\checkmark$ & 5.26  & \textbf{63.7}  & \textbf{53.9}  & 46.9  \\
\Xhline{1.0pt}
\end{tabular}
}
\vspace{-5.0pt}
\end{table*}